\documentclass[times,preprint,authoryear]{elsarticle}

\usepackage{ycviu}
\usepackage{framed,multirow}

\usepackage{amssymb}
\usepackage{amsmath}
\usepackage{latexsym}
\usepackage{dsfont}
\usepackage[ruled,linesnumbered]{algorithm2e}
\usepackage{booktabs}
\usepackage{tabularray}
\usepackage[T1]{fontenc}
\usepackage{fontawesome}
\usepackage{pdflscape}
\usepackage{fancyhdr}
\usepackage{eso-pic}

\usepackage{url}
\usepackage{xcolor}
\usepackage{tikz}
\definecolor{newcolor}{rgb}{.8,.349,.1}

\journal{Computer Vision and Image Understanding}

\newcommand{\pr}{$\mathcal{P}_{\mathcal{R}}$}
\newcommand{\pg}{$\mathcal{P}_{\mathcal{G}}$}
\newtheorem{theorem}{Theorem}

\begin{document}

\thispagestyle{empty}

\clearpage
\thispagestyle{empty}
\ifpreprint
  \vspace*{-1pc}
\fi

\thispagestyle{empty}

\ifpreprint
  \vspace*{-1pc}
\else
\fi










\ifpreprint
  \setcounter{page}{1}
\else
  \setcounter{page}{1}
\fi

\begin{frontmatter}

\title{Establishing a Unified Evaluation Framework for Human Motion Generation: A Comparative Analysis of Metrics}

\author[1]{Ali \snm{Ismail-Fawaz}\corref{cor1}} 
\cortext[cor1]{Corresponding author: 
  \textit{website}: \texttt{hadifawaz1999.githubt.io}
  }
\ead{ali-el-hadi.ismail-fawaz@uha.fr}
\author[1]{Maxime \snm{Devanne}}
\author[2]{Stefano \snm{Berretti}}
\author[1]{Jonathan \snm{Weber}}
\author[1,3]{Germain \snm{Forestier}}

\address[1]{IRIMAS Universite de Haute-Alsace, Mulhouse France}
\address[2]{MICC, University of Florence, Florence, Italy}
\address[3]{DSAI Monash University, Melbourne Australia}

\received{1 May 2013}
\finalform{10 May 2013}
\accepted{13 May 2013}
\availableonline{15 May 2013}
\communicated{S. Sarkar}

\footnotetext[1]{\textbf{This paper is under consideration at Computer Vision and Image Understanding.}}

\begin{abstract}
%
The development of generative artificial intelligence for human motion generation has expanded rapidly, necessitating a unified evaluation framework. This paper presents a detailed review of eight evaluation metrics for human motion generation, highlighting their unique features and shortcomings. We propose standardized practices through a unified evaluation setup to facilitate consistent model comparisons. Additionally, we introduce a novel metric that assesses diversity in temporal distortion by analyzing warping diversity, thereby enhancing the evaluation of temporal data. We also conduct experimental analyses of three generative models using a publicly available dataset, offering insights into the interpretation of each metric in specific case scenarios. Our goal is to offer a clear, user-friendly evaluation framework for newcomers, complemented by publicly accessible code.
\end{abstract}


\end{frontmatter}

\section{Introduction}
Evaluating generative models is one of the most challenging tasks to achieve~\citep{density-coverage-paper}.
This kind of challenge is largely absent in discriminative models, where evaluation primarily involves comparison with ground truth data.
However, for generative models, evaluation involves quantifying the validity between real samples and those generated by the model.
A common method for evaluating generative models is through human judgment metrics, such as Mean Opinion Scores (MOS)~\citep{mos-paper}.
However, this type of evaluation assumes a uniform perception among users regarding what constitutes ideal and realistic generation, which is often not the case.
For this reason, generative models require quantitative evaluation based on measures of validity between real and generated samples.
This similarity is quantified on two dimensions: fidelity and diversity.
On the one hand, fidelity is the measure of similarity between real and generated spaces on the marginal distribution scale.
On the other hand, diversity is the measure of how varied a set of samples is, indicating the extent to which the diversity of the generated set in generative models aligns with the diversity of the real set.

Generative models are used for almost every type of data ranging from images~\citep{ddpm} to time series data~\citep{time-vqvae,shapedba-paper}.
In this work, we investigate human motion generation~\citep{action2motion}, a subject that has attracted substantial interest in the research community.
This surge in interest is attributable to the availability of datasets generated by advanced technologies like Kinect cameras and motion capture systems~\citep{kinect-survey, kinecy-sensors, mo-cap}.
With these technologies, a variety of 3D human motion skeleton-based datasets have been developed.
The availability of these datasets, coupled with their importance in applications ranging from the cinematic universe to the medical field, has led the research community to take a keen interest in generating 3D skeleton-based motion samples.
An early work that addressed this new approach of handling a human motion generation model is Action2Motion~\citep{action2motion}.
It is a recurrent based deep learning model with a Conditional Variational Auto-Encoder (CVAE)~\citep{vae-paper,cvae} architecture where, for each frame, the model learns the parameters of a CVAE.
The authors in~\citep{cs-vae} proposed CS-VAE, a novel CVAE model that utilizes the first part of the sequence as the condition to encode, which is then used to generate the rest of the sequence.
With the rise of self-attention and Transformer models~\citep{transformers}, a non-recurrent CVAE with a Transformer backbone architecture, called ACTOR, is proposed in~\citep{actor}.
Further, with the significant improve of performance in language models thanks to Generative Pre-trained Transformers (GPTs)~\citep{gpt1}, the authors in~\citep{pose-gpt} proposed PoseGPT, a Quantized CVAE~\citep{q-vae}, with a GPT used on the quantized latent space to predict the next token sequentially.
More recently, the authors in~\citep{um-cvae} proposed an uncoupled CVAE (UM-CVAE) model with two encoders, one being action agnostic and the other action aware.
By considering how much Convolutional Neural Networks (CNNs) dominate the classification field for time series~\citep{cf4tsc-paper,lite-paper,pretext-task-paper,trilite-paper}, the authors in~\citep{svae-paper} showcase their contribution to human motion generation by proposing a CNN based VAE, trained in a supervised manner to classify the latent space (SVAE).
Finally, Denoising Diffusion Probabilistic Models (DDPM) have changed the course of image generation~\citep{ddpm,stable-ddpm,cold-ddpm}, and started to give effect to human motion generation as well.
For instance, in~\citep{motion-diffuse} the first DDPM dedicated for 3D skeleton based human motion generation was proposed that outperformed the state-of-the-art. 
In~\citep{phys-ddpm}, the human motion DDPM was adjusted so that it can take into consideration the physical aspects of the human skeleton.

\definecolor{mygreen}{RGB}{92, 190, 132}
\definecolor{myblue}{RGB}{92, 168, 212}

\begin{figure}
    \centering
    \includegraphics[width=0.5\textwidth]{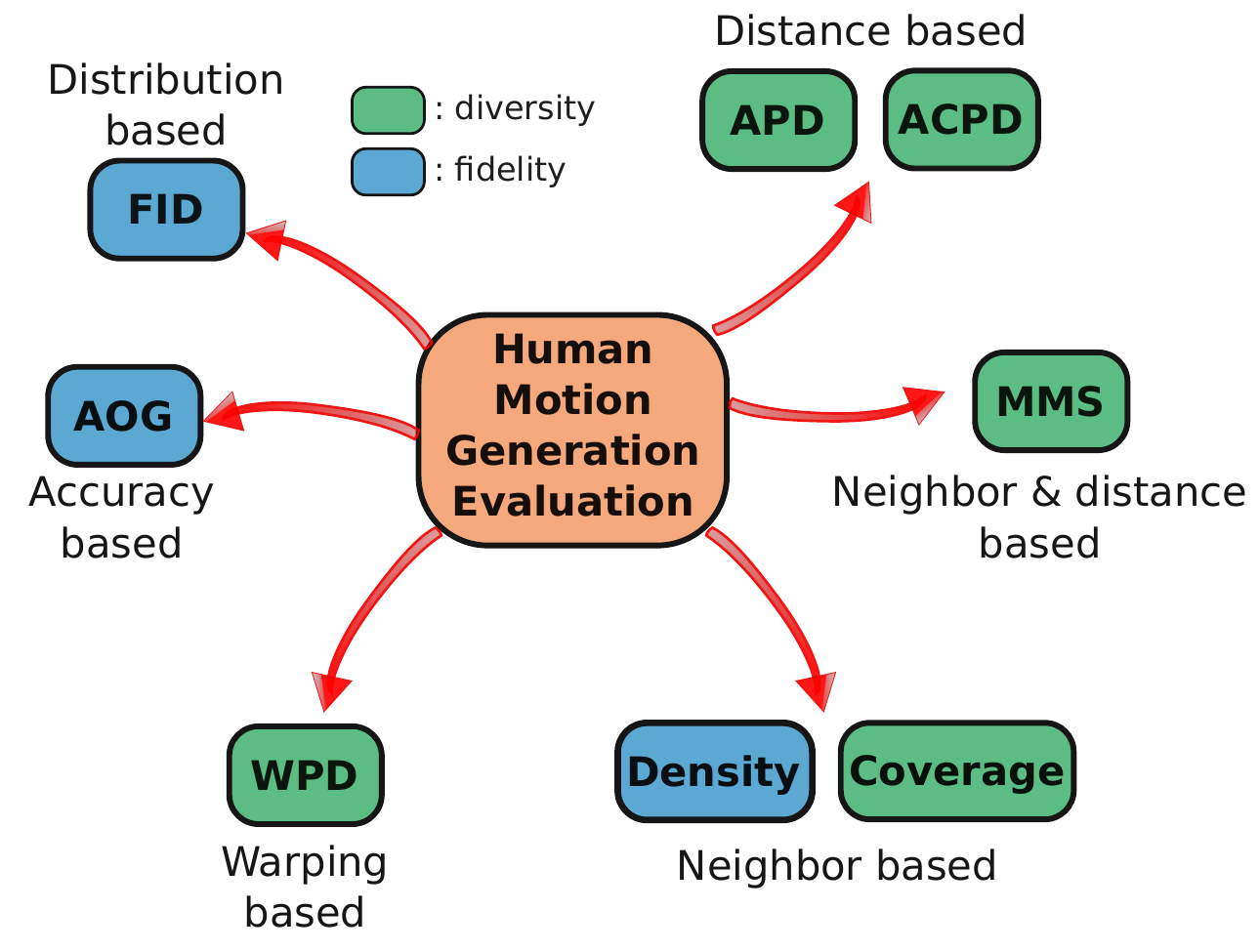}
    \caption{Summary of all the evaluation metrics for human motion generation used in this work.
    First, the metrics are divided into two groups: the fidelity metrics in \colorbox{myblue}{blue}, and diversity metrics in \colorbox{mygreen}{green}.
    Second, the metrics are divided into sub-groups indicating the measure is based on which criteria, such as FID being a distribution based metric.}
    \label{fig:metrics-summary}
\end{figure}

All the related works on human motion generation mentioned above use some common evaluation approach, while some propose their own new metric as well.
The primary challenge lies in defining the metrics for both fidelity and diversity, given that there is no singular optimal solution, \emph{no metric to rule them all}. 
For this reason, several approaches were proposed to address this issue, and novel metrics were defined as well.
Evaluating existing metrics alone is not a viable solution. Consequently, numerous diverse metrics have emerged, each tailored to measure specific aspects of fidelity or diversity.
However, the generation setups of all these previous works are not always detailed and well explained. Moreover each paper can have a different generation framework posterior to training, which can make the comparison between models somehow problematic.

We thus believe, that it is important to unify the evaluation process.
In this work, we first summarize the evaluation metrics from the literature with a unified evaluation framework in order to have a comparison that is as fair as possible.
We believe that this work can be a good starting point for newcomers in the field.
It is important to note, that through our experimental setup and analysis, we showcase that it is difficult to find the \emph{best} model that wins over all metrics, as a slight change in the architecture and hyper parameters can drastically change the metric values resulting in a different interpretation.
Instead, we consider a detailed experiment using three variants of a CVAE model trained on the same dataset, highlighting the analysis of each metric.

One special criteria for human motion data, as well as any sequential data, is the temporal dependency between its features~\citep{ucr-archive}.
A common information that has been of interest to many researchers in the field of time series is temporal distortion.
Temporal distortion is the measure of time shift, frequency change and warping between a pair of time series samples.
This is directly applied in the case of human motion sequences given they are multivariate time series~\citep{uea-archive}.
The metrics utilized in this study are not limited to human motion generation but are applicable to various generative models across different domains.
In a case where a generative model is not able to generate the same type of temporal distortions as in the real set of samples, the existing metrics do not capture this lack of diversity given they are computed on a latent representation independent of the temporal aspect of the sequences.
For this reason, additionally to presenting the existing metrics in the literature, in this work we propose a novel metric which is specific for temporal data used in the application of human motion generation.
This metric, that we called \textit{Warping Path Diversity} ($WPD$), evaluates the diversity of temporal distortions in both the real and generated space. In particular, if it is important to generate sequences with different temporal distortion such as different frequencies, then the generative model scores a high value on the $WPD$ metric.

\begin{figure*}
    \centering
    \includegraphics[width=0.9\textwidth]{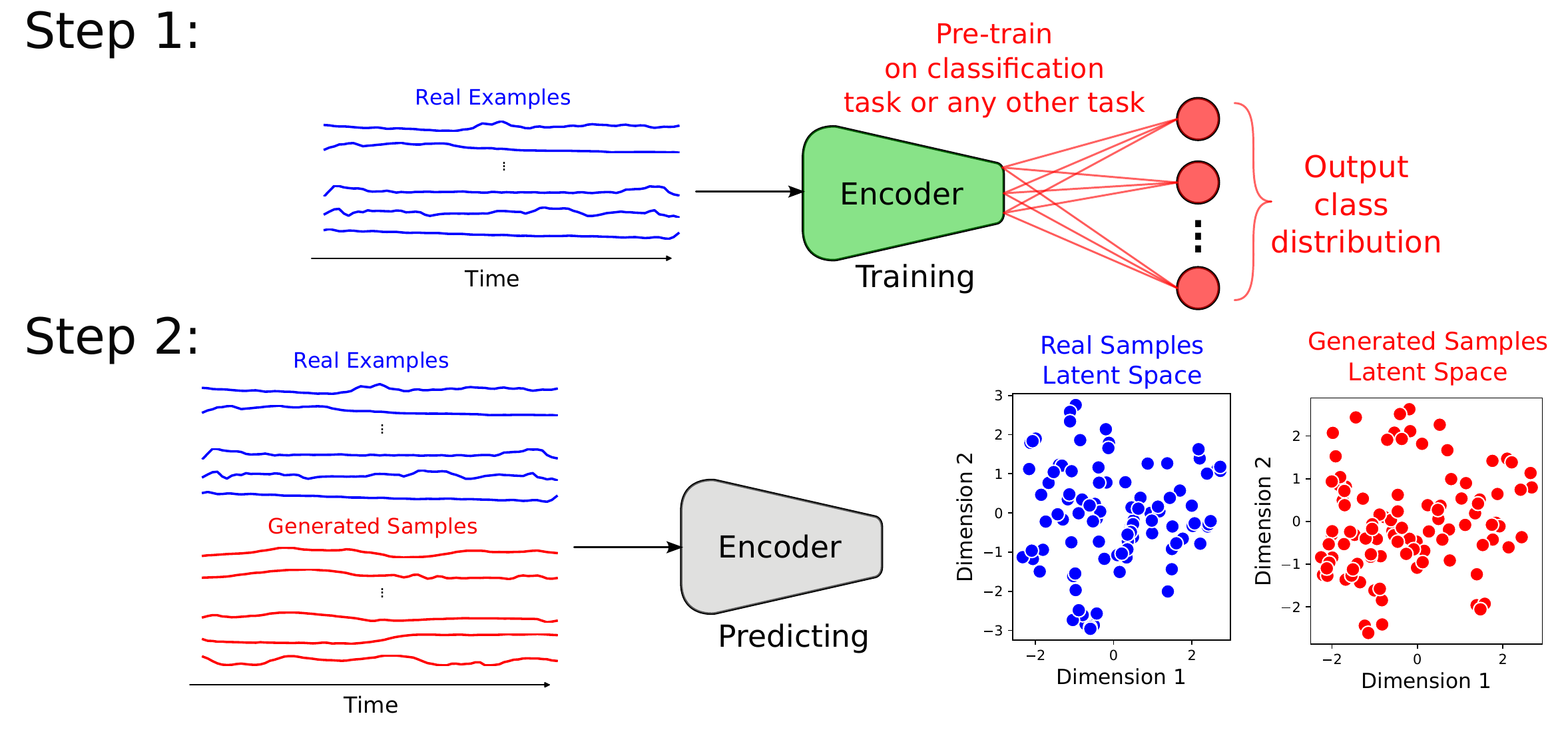}
    \caption{Two steps are followed prior to calculating evaluation measures.
    In the first step, a model is trained on some supervised task using the real set of data (in \textcolor{blue}{blue}).
    This model consists of a feature extractor encoder (in \textcolor{green}{green}) and a final layer for the supervised task.
    In the second step, the pre-trained encoder's latent representation of the real (in \textcolor{blue}{blue}) is extracted, and samples are generated (in \textcolor{red}{red}).
    The metrics are then computed over this latent representation.}
    \label{fig:latent-space}
\end{figure*}
In this work, we did not rely on data augmentation as an evaluation metric.
This is due to the fact that if we suppose there are not enough data to train a supervised model so we rely on augmented samples provided by a generative model, then the question is: how were we able to train a generative model if there are not enough data ?
Leading into a paradox problematic, we decided not to experiment with this kind of metric, instead rely on similarity between real and generated distributions through different approaches.
A brief summary of all the metrics used in this work is presented in Figure~\ref{fig:metrics-summary}.
In this figure, metrics are divided into two categories: fidelity (in \colorbox{myblue}{blue}) and diversity (in \colorbox{mygreen}{green}).
Subsequently, the metrics are then organized in sub-categories depending on which approach they utilize for the evaluation, this includes: accuracy based, distribution based, distance based, neighbor based, neighbor/distance based and warping based.

The contributions of this work are summarized as follows:
\begin{itemize}
    \item We review most of the frequently used metrics in the literature of human motion generation with their mathematical setup and their interpretation;
    \item We propose a novel metric specific to human motion generation called Warping Path Diversity ($WPD$) to measure diversity in terms of temporal distortion;
    \item We define a unified experimental setup in order to fairly evaluate and compare generative models;
    \item We support this work with a user-friendly publicly available code to evaluate new generative models on all of the considered metrics: \url{https://github.com/MSD-IRIMAS/Evaluating-HMG}.
\end{itemize}

\section{Generative Models Metrics}
The generative models evaluation metrics are divided into two sub-categories called \emph{fidelity} and \emph{diversity}.
The fidelity metrics measure the extent to which we can trust the generated samples to accurately replicate the characteristics of the real distribution.
The higher the fidelity of generated examples, the more challenging it is to differentiate between real and generated samples, resulting in a more reliable generative model.
The diversity metrics quantify the extent to which the generated samples differ from each other.
A greater diversity within a generated set indicates that the generative model's capacity to produce samples is not confined to a particular segment of the real distribution.

In what follows we first define essential terms to comprehend the rest of this work. Second, we review and detail main metrics employed in the literature for both \emph{fidelity} and \emph{diversity}.

\subsection{Definitions}\label{sec:definitions}
To understand the metrics that follow, it's necessary to establish some definitions:
\begin{itemize}
    \item A human motion sequence is referred to as $\textbf{x} \in \mathbb{R}^{D\times T}$, where $T$ corresponds to the length of the sequence and $D$ corresponds to the number of characteristics representing a human skeleton. 
    \item A set of $N$ real samples is referred to as $\mathcal{X}=\{\textbf{x}_i\}_{i=1}^{N}$ and follows a distribution \pr;
    \item A set of $M$ generated samples is referred to as $\hat{\mathcal{X}}=\{\hat{\textbf{x}}_j\}_{j=1}^{M}$ and follows a distribution \pg;
    \item A pre-trained deep learning model $\mathcal{G}\circ\mathcal{F}(.)$ is made of a feature extractor $\mathcal{F}$ and a last layer $\mathcal{G}$ achieving the desired task (e.g. classification).
\end{itemize}

\begin{figure*}[t]
    \centering
    \includegraphics[width=0.8\textwidth]{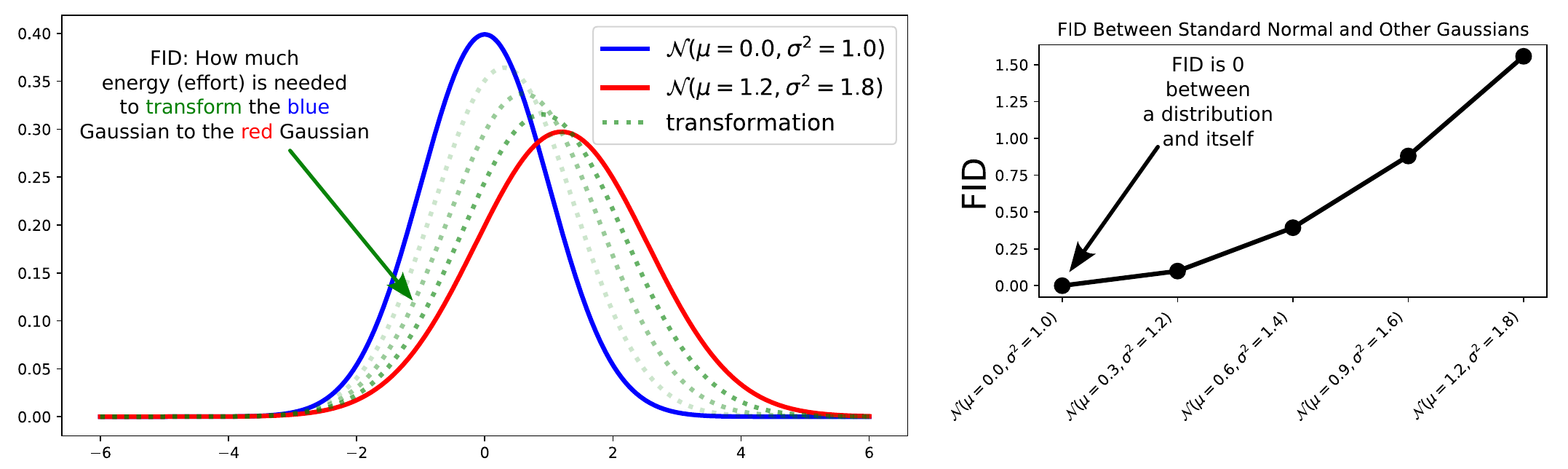}
    \caption{On the left, an example computation of the amount of energy needed ($FID$) to transform the standard Gaussian distribution (\textcolor{blue}{in blue}) to another Gaussian distribution with higher values of $\mu$ and $\sigma^2$ (\textcolor{red}{in red}).
    On the right, the amount of energy ($FID$) needed for this transformation, gradually increasing with the increase of $\mu$ and $\sigma^2$, can be observed.}
    \label{fig:fid-vs-gauss}
\end{figure*}

In order to compute most of the metrics, the deep learning model $\mathcal{G}\circ\mathcal{F}$ is trained to solve a specific task on the real set of data. This task is usually classification, hence $\mathcal{G}$ represents a \emph{softmax} layer.
This feature extractor is subsequently employed to project the real data $\mathcal{X}$ and the generated data $\hat{\mathcal{X}}$ into a latent representation.
Thus, this allows the calculation of metrics within this latent space.
This procedure is illustrated in Figure~\ref{fig:latent-space}, showcasing two steps: (1) the training a deep learning model to solve a task on the real data; and (2) the use of the feature extractor part (discarding the last task layer) to encode real and generated samples into a latent space.
For the rest of this work, we refer to the latent space of $\mathcal{X}$ and $\hat{\mathcal{X}}$ as $\textbf{V}$ and $\hat{\textbf{V}}$ respectively, such as:
\begin{equation}
    \textbf{V} = \mathcal{F}(\mathcal{X}) \;\; \text{and} \;\; \hat{\textbf{V}} = \mathcal{F}(\hat{\mathcal{X}}), 
\end{equation}

\noindent where both $\textbf{V}$ and $\hat{\textbf{V}}$ are now two-dimensional matrices, with the same number of examples of both $\mathcal{X}$ and $\hat{\mathcal{X}}$ respectively, and each dimension $f$.

For every metric detailed in this section, two versions will be computed: one is calculated using the generated samples (coupled with real samples depending on the metric), and one is calculated on the real samples.
The reason for this setup is to have a reference value of the metric over \pr.
This is done by randomly splitting $\textbf{V}$ into two subsets $\textbf{V}_1$ and $\textbf{V}_2$, and calculating the metrics as if $\textbf{V}_1$ is the latent space of the real samples and $\textbf{V}_2$ the latent space of the generated samples.

In what follows, we denote by $\mathcal{N}(\mu, \Sigma)$ a multidimensional Gaussian distribution of mean vector $\mu$ and covariance matrix $\Sigma$.

\subsection{Fidelity Metrics}
The \emph{fidelity} metrics in the literature are either distribution based, neighbor based or accuracy based.

\subsubsection{Fréchet Inception Distance ($FID$)}\label{sec:fid}
First proposed in 2017 by~\citep{fid-paper}, the $FID$ metric is still the most used measure to evaluate generative models.
It is an improvement of a previous measure, the Inception Score (IS)~\citep{is-paper}, which mainly evaluates the classification of a pre-trained Inception image classifier on the generated samples.
The Inception model is pre-trained on real samples, which means the IS measure takes into consideration, indirectly, the real distribution $\mathcal{P}_{\mathcal{R}}$.
However, it never quantifies the difference between both \pr~and~\pg. For this reason, the authors in~\citep{fid-paper} proposed a novel evaluation metric called $FID$, which is based on calculating the Fréchet Distance~\citep{fd-paper}
between two Gaussian distributions in Inception's latent space of both $\mathcal{X}$ and $\hat{\mathcal{X}}$.

The Fréchet Distance (FD)~\citep{fd-paper} measures a distance between continuous finite curves. To understand what FD measures, a famous example goes as follows:
\textit{Imagine a person and their dog, each wanting to traverse a different finite curved path. The speed of the person and the dog can vary but they are not allowed to go backward on the path. The FD between these two curves is the length of a leash, small enough so that both the person and the dog can traverse the whole finite curve.}
In the case of two probability distributions~\citep{fd-dist-paper}, the FD is calculated between the Cumulative Distribution Functions (CDFs) of both distributions.
For the case of multidimensional Gaussian distributions~\citep{fid-gaus-paper} $\mathcal{P}_1 \sim \mathcal{N}(\mu_1,\Sigma_1)$ and $\mathcal{P}_2\sim\mathcal{N}(\mu_2,\Sigma_2)$, both of dimension $f$, the FD is calculated as follows:
\begin{equation}\label{equ:fid}
    FD(\mathcal{P}_1,\mathcal{P}_2)^2 = \textit{trace}(\Sigma_1+\Sigma_2-2(\Sigma_1.\Sigma_2)^{1/2}) + \sum_{i=1}^{f}(\mu_{1,i}-\mu_{2,i})^2 ,
\end{equation}

\noindent where the values of $FD$ (or $FID$) range from $0$ to $+\infty$.

\paragraph{Setup for generative models}
\textbf{First}, the mean vectors $\mu_{f}$ and $\hat{\mu}_{f}$ of both $\textbf{V}$ and $\hat{\textbf{V}}$, respectively, are estimated empirically, as well as their covariance matrices $\Sigma_{f}$ and $\hat{\Sigma}_{f}$.
\textbf{Second}, the FD is calculated using Eq.~\ref{equ:fid}.
For consistency with the literature, we refer in the rest of this work for this metric as the $FID$ instead of FD, even though the \textbf{I}nception network is not used in the case of human motion.

\paragraph{Interpretation}
The $FID$ can be considered as the amount of energy (or effort) that is needed to transform one Gaussian distribution to another one.
This can be seen in Figure~\ref{fig:fid-vs-gauss}, where the probability density function of two Gaussian distributions is plotted (left side).
The amount of energy ($FID$) needed to transform the $\mathcal{N}(\mu=0.0, \sigma^2=1.0)$ to $\mathcal{N}(\mu=1.2, \sigma^2=1.8)$ increases when the mean and variance of the first distribution are increased (right side).
Given that no amount of energy is needed to transform a distribution to itself, the starting point of the plot in Figure~\ref{fig:fid-vs-gauss} (right side) is $0.0$. This last interpretation is consistent with the original definition of the $FID$, as it is a \emph{distance}.

\begin{figure*}[t]
    \centering    \includegraphics[width=0.8\textwidth]{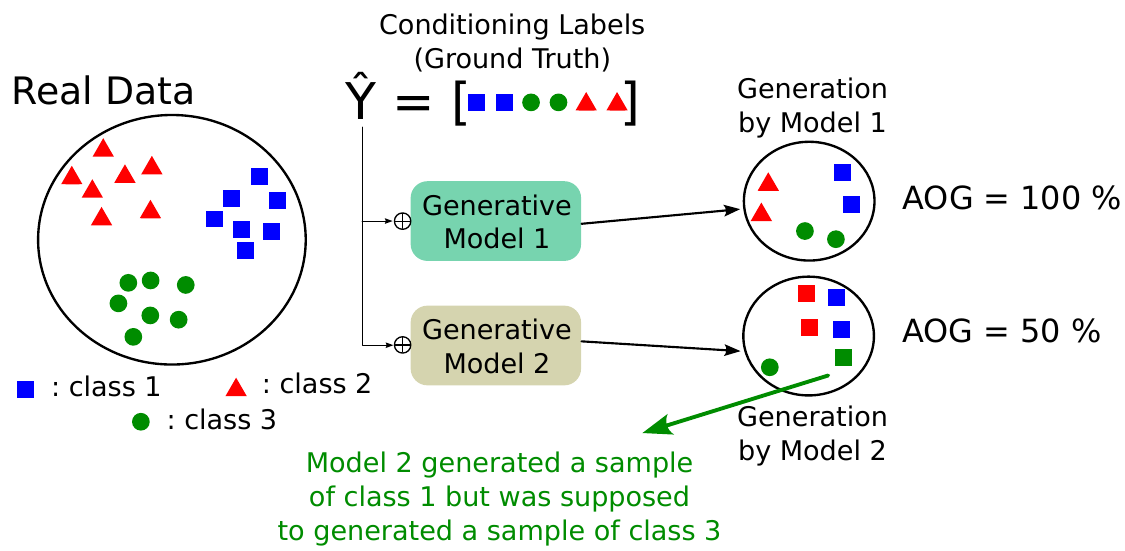}
    \caption{This example showcases the computation of the $AOG$ metric for two generative models.
    Given a real space of data, where the samples are spread over three sets of classes (\textcolor{red}{red triangles}, \textcolor{blue}{blue squares}, and \textcolor{green}{green circles}), the generative model should be able to conditionally generate new samples.
    This condition is simply to which of the possible classes the generated example should belong to.
    From the example, two generative models should be able to generate six samples, each conditioned on a class. The set of conditions used is $\hat{Y}$ (the ground truth).
    Posterior to generating, a pre-trained model classifies the generated samples. Hence, the $AOG$ metric corresponds to the accuracy of classification between the predicted class provided by the pre-trained model and the ground truth.
    In this example, \textit{Model1} obtains the optimal $AOG$ value of $100\%$, while \textit{Model2} struggles to correctly condition its generation ($AOG$ of $50\%$.}
    \label{fig:aog}
\end{figure*}

Many works in the literature mention that a lower value of $FID_{gen}$ implies a high-fidelity of generated samples: However, this can be miss-interpreted in some cases.
For instance, if we suppose a generative model learned to copy-and-paste the real samples, this would results into a perfect alignment between \pr~and~\pg. This would produce an $FID_{gen}$ of $0$, but without generating samples that bring any new value. 
For this reason, we propose that the comparison two generative models over fidelity using the $FID$ metric should follow this Theorem:
\begin{theorem}\label{the:fid}
a generative model $Gen_1$ is considered more fidelitous than another model $Gen_2$ on the $FID$ metric if $FID_{gen1} < FID_{gen2}$ while respecting the following constraint:
\begin{equation*} 
    \forall~\epsilon>0, \hspace{1cm} FID_{gen} = FID_{real} + \epsilon
\end{equation*}
\end{theorem}
More details on this setup are given in Section~\ref{sec:setup-eval}.

\subsubsection{Accuracy On Generated (AOG)}
Generative models can be trained with a condition by incorporating information about the relevant characteristics associated with each sample. 
In the case of a labeled dataset, this information can be in the form of a discrete label (class information), a continuous value associated to a sample, or even be a text description of the sample.
Usually, conditioning the generative model offers a better control on the generated sample.
For instance, the generation of a human motion sequence could be conditioned with the action the person should perform (e.g., \emph{"running", "jumping"}, etc.).

A possible approach to evaluate the conditioning capability of the generative model, is to look at the score of a classifier pre-trained on solving the task over real samples $\mathcal{X}$, assuming the generated set $\hat{\mathcal{X}}$ is an unseen set.
The classifier used here is $\mathcal{G}\circ\mathcal{F}$ defined in Section~\ref{sec:definitions}.
In this study, we refer to this metric as the Accuracy On Generated (AOG), and it is formulated as follows:
\begin{equation}\label{equ:aog}
    AOG(\hat{\mathcal{X}}, \mathcal{G}\circ\mathcal{F}) = \dfrac{1}{M}\sum_{i=1}^{M}\mathds{1}\{\mathcal{G}\circ\mathcal{F}(\hat{\mathcal{X}}_i) == \hat{Y}_i\},
\end{equation}

\noindent
where $\hat{Y}$ represents the set of labels employed to condition the generation process, serving as ground truth labels that $\mathcal{G}\circ\mathcal{F}$ is expected to predict.
Additionally, $\mathds{1}$ denotes the indicator function defined as:
\begin{equation}
    \mathds{1}\{condition\}=
    \begin{cases}
        1 & if\text{ }condition\text{ }is\text{ }True\\
        0 & if\text{ }condition\text{ }is\text{ }False
    \end{cases}
\end{equation}
\noindent where the values of $AOG$ in Eq.~\ref{equ:aog} range from $0$ to $1$.

\paragraph{Setup for generative models}
The $AOG$ metric is calculated between the ground truth labels $\hat{Y}$ and the predictions of $\mathcal{G}\circ\mathcal{F}(.)$ using Eq.~\ref{equ:aog}.

\paragraph{Interpretation}
The value of the $AOG$ metric and the generative model's conditioning capability are directly proportional.
However, a generative model with a very low value of $AOG$ may indicate that the generative model is capable of generating samples from the same label and not the whole possible set of labels.
Additionally, a generative model with a $100\%$ $AOG$, would not certainly mean that the generated samples are fidelitous with respect to \pr.
For instance if the generated samples contains a significant amount of residual noise, but enough amount to not affect the classifier's performance, then a high value of the $AOG$ metric leads to wrong conclusions about the generation's fidelity.
A visualization example on the $AOG$'s behavior is given in Figure~\ref{fig:aog}.

\begin{figure*}[t]
    \centering
    \includegraphics[width=\textwidth]{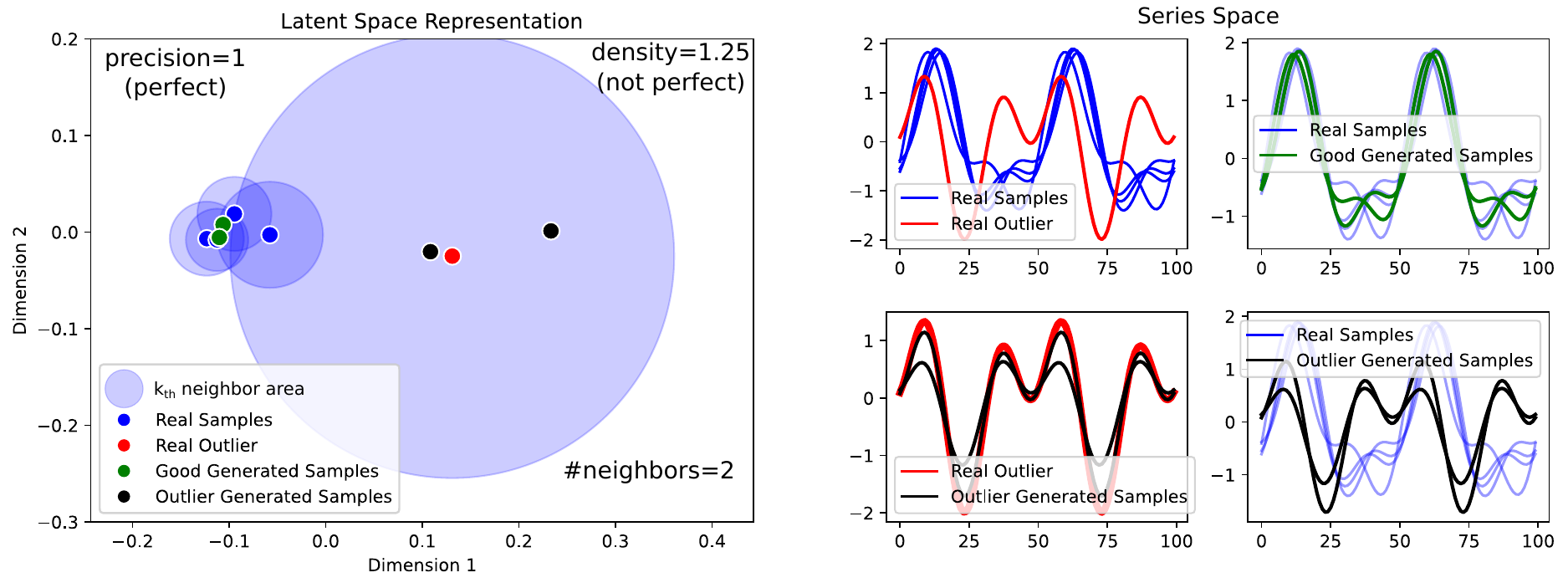}
    \caption{This example showcases the computation of both the $density$ and $precision$ metrics over a synthetic dataset.
    On the left, we present the latent representation of the real set of data (\textcolor{blue}{blue points}), the real outlier (\textcolor{red}{red point}), the generated examples around the outlier (\textcolor{black}{black points}), and generated examples around non-outliers (\textcolor{green}{in green}).
    The \textcolor{blue}{circles in blue} represent the neighborhood area of each latent point of a real sample, for both outlier and non-outlier real samples.
    On the right, we present the original series space of the data with the same associated colors.
    It can be seen that the $density$ metric overcomes the existence of an outlier and does not produce a perfect measure, $1.25 \neq 1$. However the $precision$ does not detect the outlier and produces a perfect measure of $1$.
    This example highlights how the $density$ metric concludes that the generated samples are not fully fidelitous to the real distribution because of the generations around the real outlier that a generative model should overcome.
    The number of neighbors used for both metrics is set to $2$.}
    \label{fig:density}
\end{figure*}

\subsubsection{Density and Precision}\label{sec:density}
Relying on one metric to assess the performance of generative models is not always feasible, 
especially that it is not an easy challenge to evaluate both fidelity and diversity in a single metric.
This was argued in~\citep{precision-recall-paper} that two generative models can have the same value of the $FID$ metric, but still differ in terms of fidelity and diversity, qualitatively.
Hence, the authors in~\citep{precision-recall-paper} proposed two metrics, $precision$ and $recall$ to quantitatively evaluate fidelity and diversity, respectively.
This was followed by an improved version of these two metric in~\citep{improved-precision-recall-paper} by avoiding the hypothesis of a uniformly dense latent space and estimating density functions using the $k$-nearest neighbor algorithm.
In particular, $precision$ represents the portion of samples generated by \pg~that can be sampled from \pr~as well.
The formulation of the $improved-precision$ (referred to as $precision$ in the rest of this work) is as follows:
\begin{equation}\label{equ:precision}
precision = \frac{1}{M}\sum_{j=1}^{M}\mathds{1}(\hat{\textbf{V}}_j\in manifold(\textbf{V}_1,\ldots,\textbf{V}_N), 
\end{equation}

\noindent where $manifold(\{a_1,a_2,\ldots,a_n\}) = \bigcup_{i=1}^{n}B(a_i,NND_k(a_i))$, $B(c,r)$ is a sphere in $\mathds{R}^{dimension(a_i)}$ of center $c$ and radius $r$, and $NND_k(a_i)$ is the distance from $a_i$ to its $k_{th}$ nearest neighbor in the set $\{a_j\}_{j=1,j\neq i}^{N}$. The values of $precision$ range from $0$ to $1$.

In~\citep{density-coverage-paper}, the authors identified some limitations of both $precision$ and $recall$ metrics. They proposed two novel metrics, $density$ and $coverage$ for fidelity and diversity, respectively.
In this section, we discuss the fidelity aspect only, and in Section~\ref{sec:coverage} the diversity aspect for $recall$ and $coverage$.
Two main limitations of the $precision$ metric were highlighted in~\citep{density-coverage-paper}.
Firstly, in the simplest case scenario where both real and generated samples follow the same distribution (\pr~and~\pg~are identical), the authors observed that no closed formulation can be found for the expected value of $precision$. 
Secondly, in the case where the real set of data $\mathcal{X}$ contains some outliers that can lead to the generation of outliers in $\hat{\mathcal{X}}$, the interpretation of the $precision$ to be considered \emph{good} can be mislead.

The mathematical formulation of the $density$ metric is as follows:
\begin{equation}\label{equ:density}
density = \frac{1}{k.M}\sum_{j=1}^{M}\sum_{i=1}^{N}\mathds{1}(\hat{\textbf{V}}_j\in B(\textbf{V}_i,NND_k(\textbf{V}_i))), 
\end{equation}

\noindent where the values of $density$ range from $0$ to $\dfrac{N}{k}$.

\begin{figure*}[t]
    \centering
    \includegraphics[width=\textwidth]{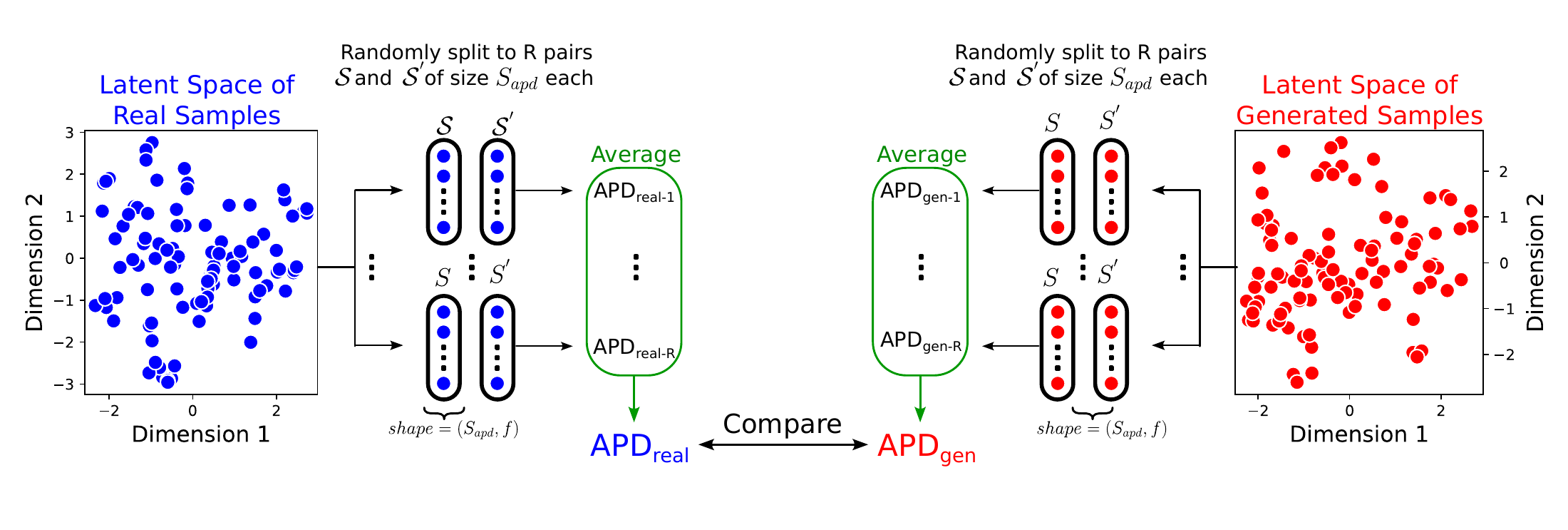}
    \caption{
    This example showcases the computation of the $APD$ metric over both real samples (\textcolor{blue}{in blue}) and generated samples (\textcolor{red}{in red}).
    For each of the latent representations of real and generated samples, two sets of randomly chosen values are constructed of size $S_{apd}$ each.
    The $APD$ metric is then calculated between these two sets of random selected values $\mathcal{S}$ and $\mathcal{S}^{'}$.
    This random selection is repeated $R$ times, the final $APD$ value is the average of all the $R$ computed $APD$ values.
    }
    \label{fig:apd}
\end{figure*}

The authors in~\citep{density-coverage-paper} proved that in the case of identical distributions \pr~and~\pg, the expected value of $density$ is $1$.
This showcases that, with enough samples and a high number of neighbors, the convergence of $density$ to the value of $1$ is proportional to the fidelity of the generated samples.
In what follows, we present a constructed example on a real human motion dataset to showcase the second benefit of density over $precision$, the detection of outlier in the real distribution.

An illustration of the outlier scenario is presented in Figure~\ref{fig:density} on a synthetically constructed dataset.
Figure~\ref{fig:density} showcases on the left the latent space representation of five real samples and four generated samples.
Among the five real samples, an outlier is present and highlighted in red in both the latent space (left) and series space (right).
It can be seen from the figure that some generated samples are correctly placed in the distribution of the four real samples, and two other generated are closer to the fifth real outlier.
The $precision$ metric, in this scenario, fails to identify the existence of an outlier in the real distribution, resulting in high $precision$ of $1$ usually interpreted as a high fidelity score.
However, the $density$ metric is able to identify that two generated samples are only present in one neighbor circle, belonging to the outlier sample, hence producing a non-perfect value $1.25 \neq 1$.

\paragraph{Setup for generative models}
\textbf{First}, we find the distance of each sample in $\mathcal{F}(\mathcal{X})$ to its $k_{th}$ nearest neighbor in $\mathcal{F}(\mathcal{X})$.
\textbf{Second}, we calculate both the $precision$ and $density$ metrics using Eq.~\ref{equ:precision} and Eq.~\ref{equ:density}.

\paragraph{Interpretation}
On the one hand, the $precision$ metric counts the number of generated samples existing in at least one neighbor sphere of any real samples.
On the other hand, $density$ counts the number of neighbor spheres containing each of the generated samples.
While both metrics quantify how fidelitous generated samples are to real samples, the $density$ metric quantifies this aspect individually between all pairs of real and generated samples.
In contrast, the $precision$ metric falls short when considering the union of all neighboring spheres, as it does not assess the bias of some generated samples due to the existence of a real outlier.

\subsection{Diversity Metrics}\label{sec:diversity}
Evaluating \emph{fidelity} measures only does not result in a full reliability on generated samples. For this reason diversity measures are required.
In what follows, we present the \emph{diversity} metrics used in the literature, ranging from distance based approaches to neighbor based approaches.

\subsubsection{Average Pair Distance ($APD$)}

Originally proposed in 2018 in~\citep{apd-paper} to compute distances between images and adapted for human motion generative models' evaluation in 2020 in~\citep{action2motion}, the Average Pair Distance ($APD$) has ever since been the go-to metric for evaluating diversity of generative models.
The $APD$ metric can evaluate the diversity of any set of data, not necessarily generated ones.
This is done through calculating the average Euclidean distance between randomly selected pairs of samples from the same dataset.
This random selection is repeated $R$ times over $S_{apd}$ number of pairs for each experiment. The final value of the $APD$ metric is the average outcome of the repeated experiments.
The $APD$ metric calculated on the generated set of samples, for one random selection $r~\in~\{1,2,\ldots,R\}$ is formulated as follows:
\begin{equation}\label{equ:apd}
    APD_r(\hat{\mathcal{X}}) = \dfrac{1}{S_{apd}}\sum_{i=1}^{S_{apd}} \sqrt{\sum_{j=1}^{f} (\mathcal{S}_{i,j} - \mathcal{S}^{'}_{i,j})^2} ,
\end{equation}

\noindent where $\mathcal{S}$ and $\mathcal{S}^{'}$ are two randomly selected subsets of $\hat{\textbf{V}}=\mathcal{F}(\hat{\mathcal{X}})$, i.e. $\mathcal{S},\mathcal{S}^{'}~\subset \hat{\textbf{V}}$.

The $APD$ metric is then calculated by averaging over the $R$ random experiments:
\begin{equation}\label{equ:apd-all}
APD(\hat{\mathcal{X}}) = \dfrac{1}{R}\sum_{r=1}^{R} APD_r(\hat{\mathcal{X}}),
\end{equation}

\noindent where $APD_r(\hat{\mathcal{X}})$ is calculated using Eq.~(\ref{equ:apd}).
An illustration of the $APD$ metric is represented in Figure~\ref{fig:apd} showcasing the same procedure followed to calculate both $APD_{real}$ and $APD_{gen}$.

\paragraph{Setup for generative models}
We calculate the $APD$ metric using Eq.~\ref{equ:apd}~and~\ref{equ:apd-all}.

\paragraph{Interpretation}
The $APD$ metric evaluates if a generative model is able to overcome the model collapse
phenomena that leads to generating the same outcome over and over.
If the expected outcome of the generative model is to be as diverse as possible, then the $APD$ metric should be as high as possible.
However, an issue may occur with this interpretation: What happens in the case where $APD_{gen} > APD_{real}$ ?
For this reason, as interpreted in~\citep{action2motion}, we present the following theorem:
\begin{theorem}\label{the:apd}
a generative model $Gen_1$ is considered more diverse than another model $Gen_2$ if $APD_{gen1} > APD_{gen2}$, while respecting the following constraint:
\begin{equation*} 
    \forall~\epsilon>0, |APD_{gen1} - APD_{real}| < \epsilon .
\end{equation*}
\end{theorem}
In simpler words, a generative model should not exceed the limit in terms of $APD$ diversity of the real distribution \pr.
Proving Theorem~\ref{the:apd} can be done by using a counter example to break its negation.
For instance, if the real data has a diversity of $APD_{real}=5$, and the generative model is not trained, but randomly initialized and used to generated samples, then 
the diversity of generated samples $APD_{gen}$ can be higher than $5$, while the generations are simply random, and without any correlation with \pr~exists.

\begin{figure*}[t]
    \centering
    \includegraphics[width=\textwidth]{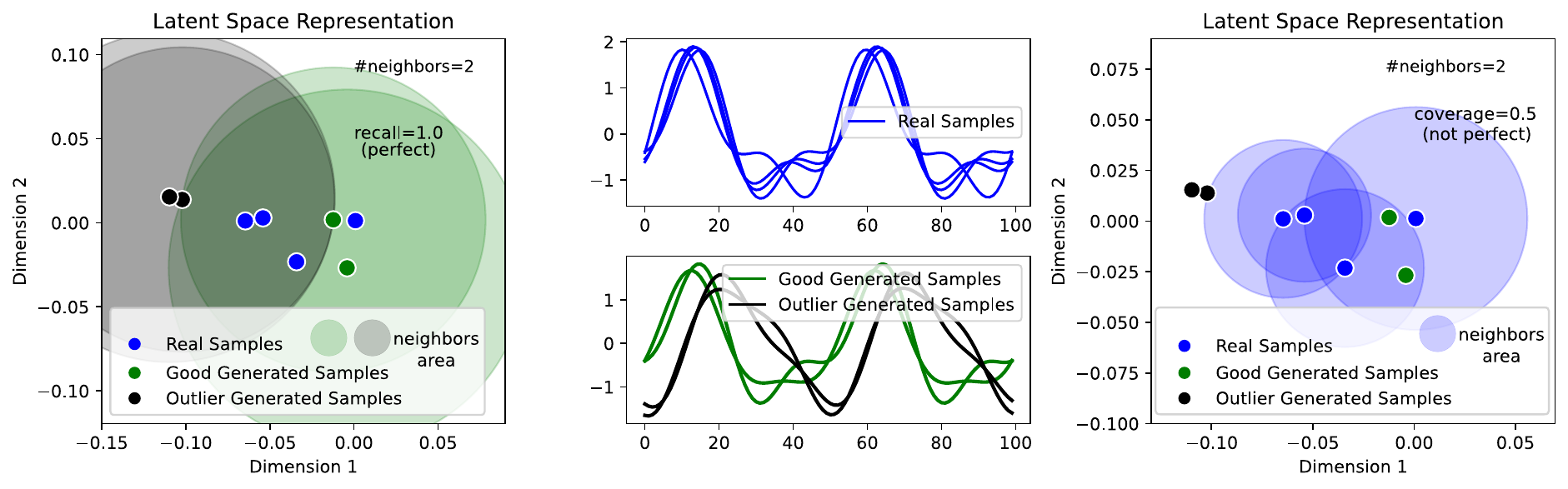}
    \caption{Computation of both the $covarage$ and $recall$ metrics over a synthetic dataset.
    In the left side and right side of the figure (same plot with different interpretation), we present the latent space representation of real samples (\textcolor{blue}{in blue}) and generated samples (\textcolor{green}{in green} and in black).
    The points \textcolor{green}{in green} represent generated samples that are similar to real samples \textcolor{blue}{in blue}.
    The points in black represent outliers in the generated space.
    In the middle part of the figure, we show the original series space of the data with the same associated colors, where it highlights that the outlier generations (series in black) are not similar to the original space of series \textcolor{blue}{in blue} unlike the reliable generated samples \textcolor{green}{in green}.
    The left plot of the latent space contains the neighbor areas around the generated samples, as the $recall$ metric requires, and the right plot of the latent space contains the neighbor areas around the real samples as the $coverage$ metric requires.
    It can be seen that the $coverage$ metric overcomes the existence of outlier generations and produces a non-perfect measure of $0.5 < 1-1/4$ in contrary to the $recall$ metric that produces a perfect measure of $1$.
    This example highlights how the $coverage$ metric concludes that the diversity in the generated space is not as good as the real space because of the existence of generated outliers.
    The number of neighbors used for both metrics is set to $2$.}
    \label{fig:coverage}
\end{figure*}

\subsubsection{Average per Class Pair Distance ($ACPD$)}
Similar to the $APD$, the Average per Class Pair Distance ($ACPD$) evaluates the diversity of the generated samples, with the same interpretation as for the $APD$.
The two metrics differ in terms of which scale is diversity evaluated.
For instance, the $APD$ metric evaluates on the scale of \pr, but it never evaluates the diversity on the scale of each sub-cluster in \pr.
For this reason, the $ACPD$ metric calculates the average $APD$ on each sub-cluster in \pr.
The mathematical formulation of the $ACPD$ is presented in the following:
\begin{equation}\label{equ:acpd}
ACPD(\hat{\mathcal{X}}) = \dfrac{1}{C.S_{acpd}}\sum_{i=1}^{S_{acpd}}\sqrt{\sum_{j=1}^{f} (\mathcal{S}_{c,i,j} - \mathcal{S}^{'}_{c,i,j})^2} ,
\end{equation}

\noindent where $C$ is the total number of classes in \pr, $\mathcal{S}_c,\mathcal{S}^{'}_c$ are randomly selected subsets from $\hat{\textbf{V}}[\hat{Y} == c]$, and $\hat{Y}$ are the labels used to generated $\hat{\mathcal{X}}$.
In other words, contrary to $APD$, the $ACPD$ metric randomly selects the subsets $\mathcal{S}$ and $\mathcal{S}^{'}$ from the latent representation $\hat{\textbf{V}}$ of $\hat{\mathcal{X}}$ that were generated using the label $c$, hence the notation $\mathcal{S}_c$ and $\mathcal{S}^{'}_c$.

Similarly to $APD$, given the randomness factor in $ACPD$, the experiment is repeated $R$ times to calculates $ACPD_r$, $r~\in~\{1,2,\ldots,R\}$, and finally averaged as:
\begin{equation}\label{equ:acpd-all}
ACPD(\hat{\mathcal{X}}) = \dfrac{1}{R}\sum_{r=1}^{R}ACPD_r(\hat{\mathcal{X}}).
\end{equation}

\noindent It is important to note that this metric is restricted only to labeled datasets.

\paragraph{Setup for generative models}
We calculate $ACPD$ using Eq.~\ref{equ:acpd}~and~\ref{equ:acpd-all}.

\paragraph{Interpretation}
The $ACPD$ metric evaluates the diversity of the generated samples per sub-cluster in \pr, hence evaluating if a model is able to overcome the generation of samples from only one cluster in \pr.
This last problem would lead to generating high diversity in one cluster (class) and less for others.
This is because machine learning often encounters imbalanced labeled data, and the $ACPD$ metric helps to assess whether a generative model can effectively address this issue.

Similarly to the $APD$ metric, $ACPD$ is calculated on both real and generated samples producing $ACPD_{real}$ and $ACPD_{gen}$, where a generative model is considered class diverse when $ACPD_{gen}$ is close to $ACPD_{real}$.

\subsubsection{Coverage and Recall}\label{sec:coverage}
As mentioned in Section~\ref{sec:density}, the authors in~\citep{density-coverage-paper} proposed two metrics to replace the improved $precision$ and $recall$ metrics~\citep{improved-precision-recall-paper,precision-recall-paper}.
The $recall$ metric represents the number of real examples sampled using \pr that can be sampled using \pg.
This is done by counting the number of real samples that exist in at least one neighborhood of a generated sample.
The $recall$ metric is formulated as:
\begin{equation}\label{equ:recall}
recall = \dfrac{1}{N}\sum_{i=1}^{N}\mathds{1}(\textbf{V}_i~\in~manifold(\hat{\textbf{V}}_1,\hat{\textbf{V}}_2,\ldots,\hat{\textbf{V}}_M)) ,
\end{equation}
\noindent where $manifold(.)$ and $\mathds{1}(.)$ follow the same definition detailed in Section~\ref{sec:density}. The $recall$ metric is bounded between $0$ and $1$.

Some limitations of the $recall$ metric have been identified by authors of~\citep{density-coverage-paper}, and can be summarized as:
\begin{enumerate}
    \item Basing the neighborhood areas on the generated samples can produce misleading interpretation given the higher chance of having an outlier sampled by \pg~compared to an outlier sampled by \pr.
    \item No closed form solution does exist for the expected value of the $recall$ metric in the case of \pr~and~\pg~being identical distributions.
\end{enumerate}

For this reason, authors of~\cite{density-coverage-paper} proposed the $coverage$ metric relying on neighborhood areas around the real samples $\mathcal{X}$.
The $coverage$ metric counts the number of real samples that include at least one generated sample from $\hat{\mathcal{X}}$.
The $coverage$ metric is formulated as:
\begin{equation}\label{equ:coverage}
coverage=\dfrac{1}{N}\sum_{i=1}^{N}\mathds{1}(\exists~j~s.t.~\hat{\textbf{V}}_j~\in~B(\textbf{V}_i,NND_k(\textbf{V}_i))) ,
\end{equation}

\noindent where $B(.,.)$ and $NND_k(.)$ follow the same definitions detailed in Section~\ref{sec:density}. The $coverage$ metric is bounded between $0$ and $1$.

The $coverage$ metric overcomes the \textbf{first} limitation of $recall$ by avoiding to use neighborhoods of generated samples.
A simple example of how $recall$ fails to detect the over-estimated neighborhood of generated samples and how $coverage$ overcomes it, is illustrated in Figure~\ref{fig:coverage}.
We utilize for this illustration the same synthetic example as in Figure~\ref{fig:density}.
On the one hand, it can be seen that the $recall$ metric results in a perfect diversity (left scatter plot).
This is due to the over-estimated neighborhood produced because of the outlier generated samples.
On the other hand, the $coverage$ metric is able to distinguish between good and outlier generated samples by simply relying on the neighborhood of the real samples.

The $coverage$ metric overcomes as well the \textbf{second} limitation of $recall$, as proven in~\citep{density-coverage-paper}.
The closed form solution of the expected value of $coverage$ in the case of identical distributions \pr~and~\pg~is simplified to:
\begin{equation}\label{equ:expected-coverage}
\mathds{E}[coverage] = 1-\dfrac{(N-1)\ldots(N-k)}{(M+N-1)\ldots(M+N-k)},
\end{equation}

\noindent which reduces to, in the case where both $N$ and $M$ are high enough:
\begin{equation}\label{equ:expected-coverage-high-nm}
\mathds{E}[coverage] = 1 - \dfrac{1}{2^k} .
\end{equation}

\paragraph{Setup for generative models}
\textbf{First}, we find the distance of each sample in $\hat{\textbf{V}}$ to its $k_{th}$ nearest neighbor in $\hat{\textbf{V}}$ and calculate the $recall$ metric using Eq~\ref{equ:recall}.
\textbf{Second}, we find the distance of each sample in $\textbf{V}$ to its $k_{th}$ nearest-neighbor in $\textbf{V}$ and calculate the $coverage$ metric using Eq.~\ref{equ:coverage}.

\paragraph{Interpretation}
On the one hand, the $recall$ metric produces the portion of real samples that falls inside at least one neighborhood of the generated samples.
On the other hand, the $coverage$ metric produces the complete opposite, which is the proportion of the real samples that contains at least one generated sample in their neighborhood.
While both metrics evaluate the diversity aspect of a generative model, the $coverage$ metric is a less risky approach relying on real neighborhoods.

Given the $coverage$ metric has an expected value of $1-\dfrac{1}{2^k}$ in the case of identical \pr~and~\pg, then the interpretation of a high diversity depends on the choice of the number of neighbors $k$.

\subsubsection{Mean Maximum Similarity (MMS)}

Originally proposed in~\citep{mms-paper}, the Mean Maximum Similarity ($MMS$) is an evaluation metric measuring novelty in the generated set.
For instance, a generative model can simply learn to generate almost exactly the same data used for training, leading to as much diversity as the real set of data.
However, this can be misleading, as such generative model is not solving the task at hand.
For this reason, the authors in~\citep{mms-paper} proposed a new metric to quantify novelty, which we interpret in this work as diversity.

\begin{figure*}
    \centering
    \includegraphics[width=\textwidth]{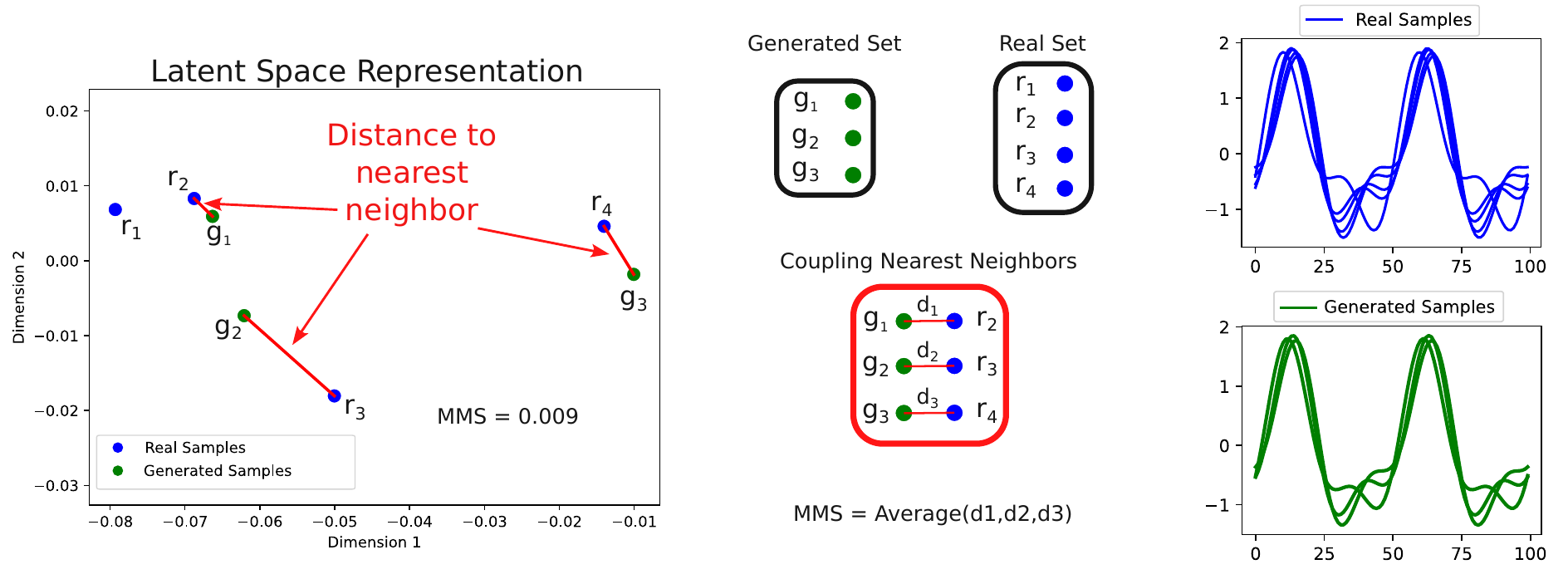}
    \caption{
    This example showcases the computation of the $MMS$ metric over a synthetic dataset.
    The left side of the figure consists of the latent representation of both real samples (\textcolor{blue}{in blue}) and generated samples (\textcolor{green}{in green}).
    The right side of the figure consists of the original series space of real and generated samples with the same associated colors.
    First, for each generated point in the latent space (left side), we find its nearest neighbor in the real set.
    Second, the Euclidean distance is calculated between each pair of generated point and its nearest neighbor.
    Third, the final $MMS$ metric is the average of all computed distances.
    }
    \label{fig:mms}
\end{figure*}

The $MMS$ quantifies novelty/diversity by averaging the distances of each of the generated samples to its real nearest samples.
It is given by:
\begin{equation}\label{equ:mms}
MMS = \dfrac{1}{M}\sum_{j=1}^{M}\sqrt{\sum_{d=1}^{f}(\hat{\textbf{V}}_{j,d} - \textbf{V}_{NN_j,d})^2} , 
\end{equation}

\noindent where $\textbf{V}_{NN_j}$ (from the real set) is the nearest neighbor to $\hat{\textbf{V}}_j$ (from the generated set).
A visual representation of the $MMS$ metric is shown in Figure~\ref{fig:mms}.

\paragraph{Setup for generative models}
\textbf{First}, for every sample in $\hat{\textbf{V}}$, the distance to its nearest neighbor in $\textbf{V}$ is calculated and averaged to produce $MMS_{gen}$.
\textbf{Second}, for every sample in $\textbf{V}$, the distance to its second nearest neighbor in $\textbf{V}$ is calculated and averaged to produce $MMS_{real}$.
We do not utilize the $\textbf{V}_1$ and $\textbf{V}_2$ sets for the $MMS$ metric, as we are following the setup of calculating $MMS_{real}$ in the original paper~\citep{mms-paper}.

\paragraph{Interpretation}
The authors in~\citep{mms-paper} interpreted their proposed $MMS$ metric as $MMS_{gen}$ should always be higher than $MMS_{real}$ to indicate high-novelty.
However, a limit for this metric is observed when the model is generating random samples far from the real set; in this case, the novelty is simply an over estimation of the generated population.
For this reason, this metric alone is not sufficient for an easy interpretation of a generative model's performance.
For this metric, as proposed in the original paper~\citep{mms-paper}, it is important to note that $MMS_{real}$ is calculated between the real set of samples and itself entirely, and not between two sub-sets as for all other metrics.

\section{Proposed Metric: Warping Path Diversity ($WPD$)}
\begin{figure*}[t]
    \centering
    \includegraphics[width=0.9\textwidth]{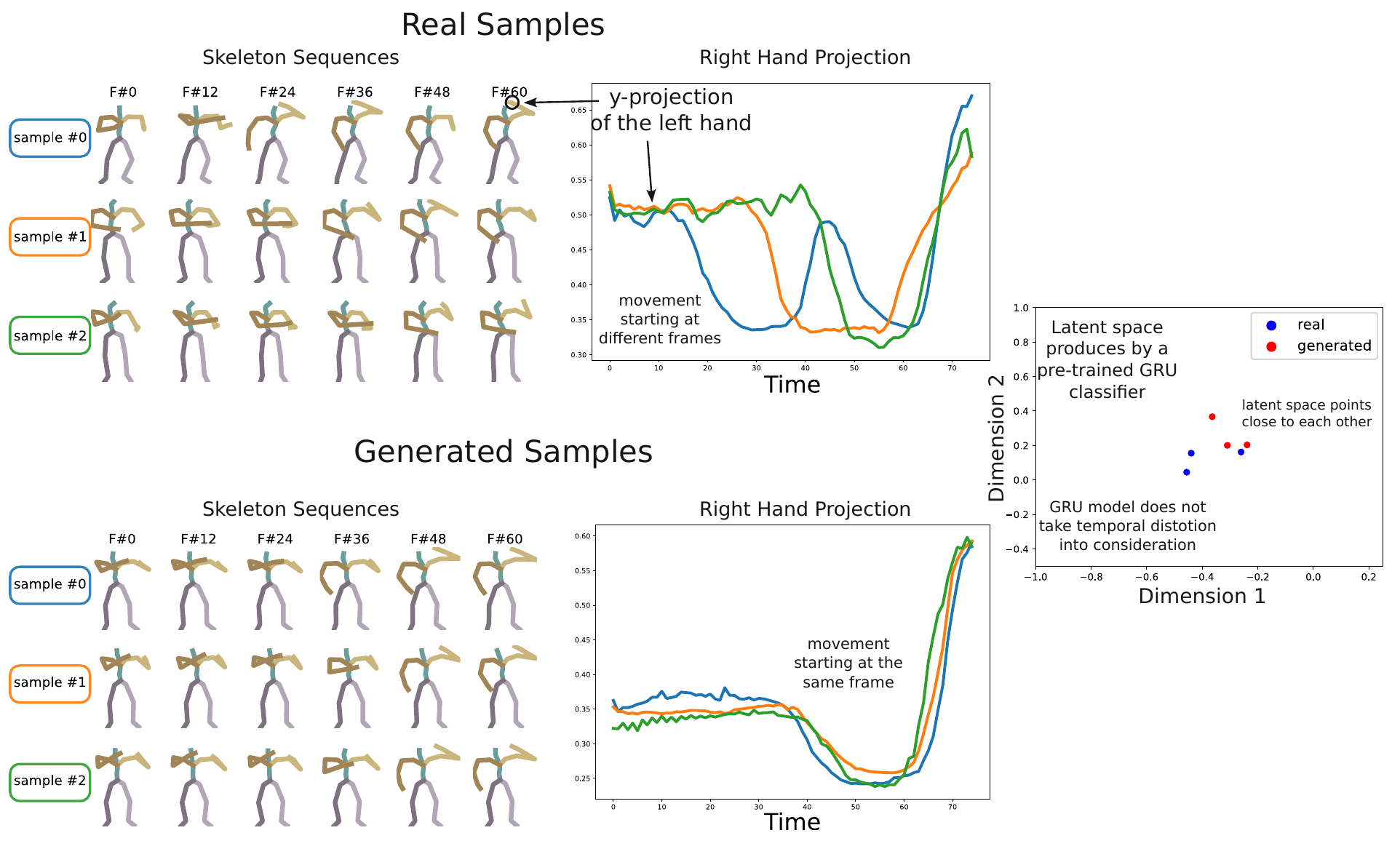}
    \caption{
    This figure illustrates the need for a temporal distortion diversity metric in certain applications. It showcases three real and three generated human motion sequences performing the "drink-with-left-hand" action. The top and bottom sections on the left display the $y-$axis projection of the subject's left hand motion as univariate time series in \textcolor{blue}{blue}, \textcolor{orange}{orange} and \textcolor{green}{green}. The real samples demonstrate variability in the starting frame for lifting the bottle, whereas the generated samples show a consistent starting frame. On the right, the latent representation of both real (\textcolor{blue}{in blue}) and generated (\textcolor{red}{in red}) samples is depicted using a pre-trained GRU classifier. This representation highlights that the pre-trained model does not account for temporal distortion diversity.}
    \label{fig:do-we-need-warping}
\end{figure*}

Every diversity metric detailed in Section~\ref{sec:diversity} utilizes a pre-trained encoder $\mathcal{F}$ to extract some latent features. This is necessary given most metrics assume an input latent space.
However, for temporal data, particularly human motion sequences, this process may introduce temporal distortions that are significant in some contexts.
Warping diversity can be caused by more than one factor such as temporal shifting, frequency change, etc.
An example considering the left-hand $y-$coordinate of skeletons is illustrated in Figure~\ref{fig:do-we-need-warping}.
It can be seen that the real samples (top of Figure~\ref{fig:do-we-need-warping}) from the HumanAct12 dataset start the action of drinking with left-hand at different frames.
However, we can see in the generated space (bottom of Figure~\ref{fig:do-we-need-warping}) that the action of drinking with left-hand always starts at the same frame, highlighting a lack of warping diversity.
As shown in the right side of Figure~\ref{fig:do-we-need-warping}) the pre-trained feature extractor $\mathcal{F}$ produces very close latent representations for all these sequences, highlighting the fact that $\mathcal{F}$ does not take into consideration the temporal distortion.
The $APD$ metric, for instance, calculates the average Euclidean distance in the latent space without taking into account the warping diversity.

For this reason, we proposed in this work a novel metric to quantify this type of diversity that relies on temporal distortions.
We utilized the Dynamic Time Warping (DTW)~\citep{dtw-music} similarity measure to extract these temporal distortions between two sequences.

In what follows, we present a background on DTW before digging into how the metric is calculated and how it can evaluate diversity in warping.

\subsection{Background on Dynamic Time Warping (DTW)}
Dynamic Time Warping~\citep{dtw-music} is a similarity measure between temporal sequences. It relies on finding the optimal temporal alignment between two sequences followed by a simple Euclidean distance on the aligned series.
The computation of DTW is presented in Algorithm~\ref{alg:dtw}.

\begin{algorithm}[t]\label{alg:dtw}
\caption{Dynamic Time Warping (DTW)}
\KwData{Sequence $x$, Sequence $y$}
\KwResult{DTW distance}
Set $L_x = len(x)$;
Set $L_y= len(y)$;
Initialize a matrix $D$ of size $L_x+1 \times L_y+1$\;
Set $D[0,0] = 0$\;
Set $D[:,0] = \infty$\;
Set $D[0,:] = \infty$\;
\For{$i \gets 1$ \KwTo $L_x$}{
    \For{$j \gets 1$ \KwTo $L_y$}{
        Set $d = (x[i]-y[j])^2$\;
        Set $D[i][j] = d + \min(D[i-1,j], D[i,j-1], D[i-1,j-1])$\;
    }
}
\Return $D[L_x,L_y]$\;
\label{dtwalgorithm}
\end{algorithm}

\begin{figure}[!ht]
    \centering
    \includegraphics[width=0.5\textwidth]{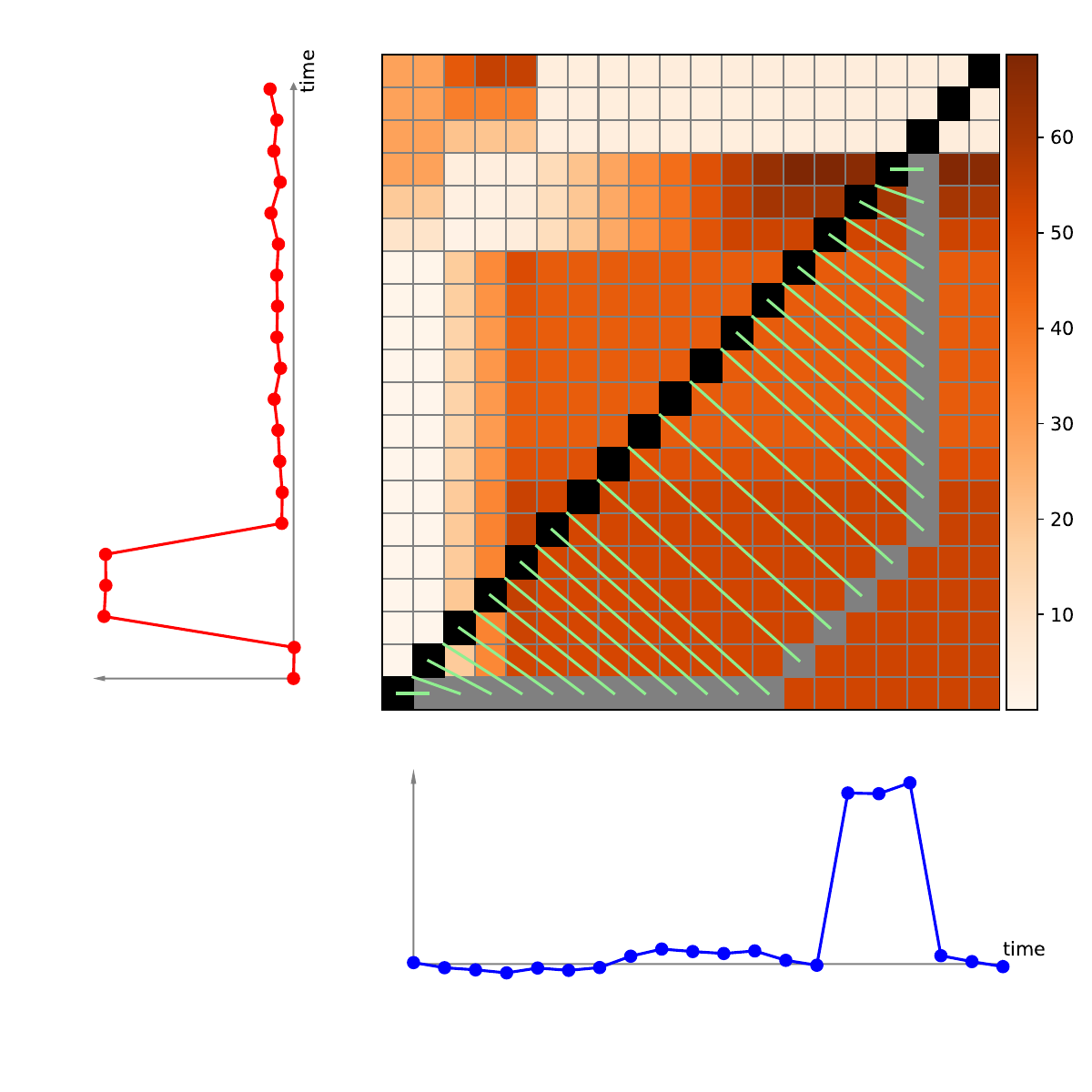}
    \caption{
    Distance matrix between two time series where each point in the heat map corresponds to the squared difference between the associated time stamps. The optimal DTW path (\textcolor{gray}{in gray}) summarizes the temporal distortion between the two time series.
    The connections (\textcolor{green}{in green}) between the warping path and the diagonal (in black), represent how far these two series are from having no temporal distortion between them.}
    \label{fig:dtw-path}
\end{figure}

In simple words, DTW finds the optimal warping path that is admissible between two sequences.
A path is a set of pairs of indices $\pi=\{(\pi_{1,1},\pi_{1,2}),(\pi_{2,1},\pi_{2,2}),\ldots,(\pi_{L_{\pi},1},\pi_{L_{\pi},2})\}$, where a pair of indices represent an alignment of these two points from $x$ and $y$.
Given two sequences, $x$ and $y$ of lengths $L_x$ and $L_y$, respectively, a warping path $\pi$ between those two sequences is considered admissible if and only if:
\begin{itemize}
    \item $\pi_0 = (0,0)$
    \item $\pi_{L_{\pi}} = (L_x,L_y)$
    \item The sequence should be monotonically increasing in both $i$ and $j$, where all indices of both time series should appear at least once, so at element $l$ of $\pi$:
    $$ i_{l-1} \leq i_l \leq i_{l-1} + 1, \; \;
    j_{l-1} \leq j_l \leq j_{l-1} + 1$$.
    \\
\end{itemize}
Figure~\ref{fig:dtw-path} illustrates the DTW matrix calculated between two sequences and the optimal warping path is presented in grey.
This optimal warping path is then used to align both series before applying the Euclidean distance between them.

\subsection{Warping Path Diversity (WPD)}
For simplicity, we consider in this section that both sequences $x$ and $y$ are of the same length $L_x=L_y=L$.
For the two sequences in Figure~\ref{fig:dtw-path}, three scenarios can occur:
\begin{enumerate}
    \item Worst case scenario, with worst alignment and the length of the path equal to $2.L$ (through matrix's edges);
    \item Best case scenario, with best alignment and the length of the path equal to the length of the diagonal. Applying DTW is equivalent to directly use Euclidean distance on $x$ and $y$;
    \item Temporal distortions are present, the path is not on the diagonal but its length is smaller than $2.L$.
\end{enumerate}

\begin{figure}[!ht]
    \centering
    \includegraphics[width=0.5\textwidth]{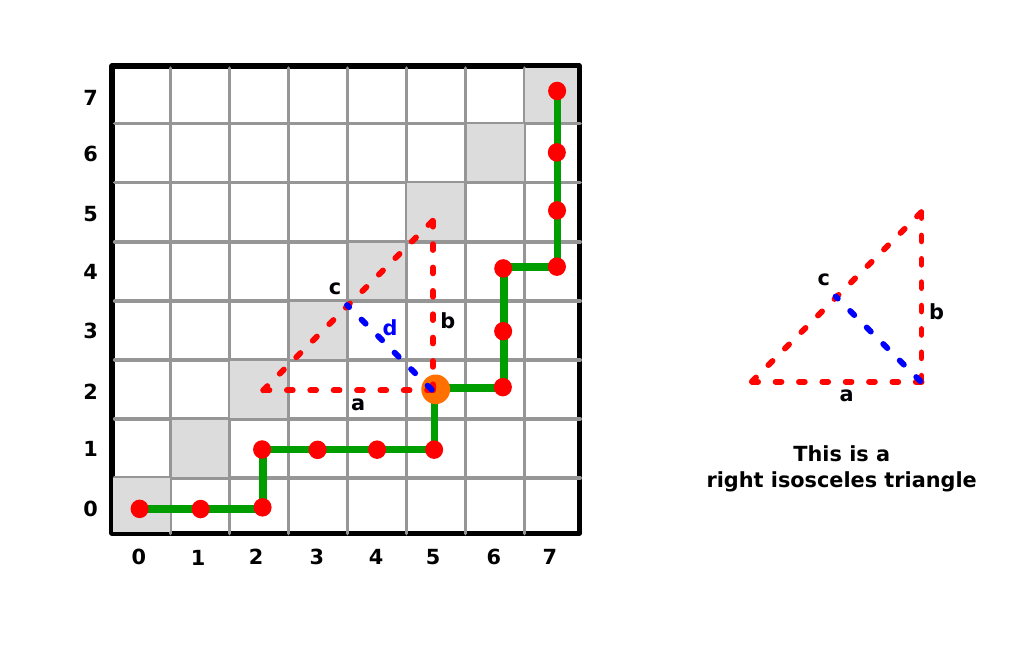}
    \caption{Basis of the mathematical formulation of the $WPD$ metric.
    For each point on the warping path, the associated triangle is always a right isosceles triangle, in the case of equal length series.}
    \label{fig:dtw-triangle}
\end{figure}

In order to measure how much diverse both sequences are in terms of temporal distortions (warping), we propose to quantify how far the warping path is from the diagonal. 
For this reason, we need to calculate the sum of distances from each point in the warping path to the diagonal.
A visual representation of this distance calculation is presented in Figure~\ref{fig:dtw-triangle}.
We consider each point on the path being in a integer coordinate space with both axis vary from $0$ to $L-1$.
Given the case of equal length sequences, the triangle presented in Figure~\ref{fig:dtw-triangle} on each point of the warping path is a right isosceles triangle.
This information leads to the hypotenuse's median being equal to half its length.
The hypotenuse's length can be simply calculated using the Pythagorean theorem.
The distance of each point on the path $\pi$ to the diagonal can be calculated as follows:
\begin{equation}\label{equ:distance-to-diagonal}
distance(\pi_k,diagonal) = \dfrac{\sqrt{2}}{2}|\pi_{k,1}-\pi_{k,2}| .
\end{equation}

\paragraph{Proof}
Using the annotations of $a,~b,~c$ and $d$ in Figure~\ref{fig:dtw-triangle} as the summit of points $\pi_k$ on the warping path $\pi$, we present the following formulation of $distance(\pi_k,diagonal)$:
\begin{equation*}
    \begin{split}
        d &= \dfrac{1}{2}\sqrt{c^2} = \dfrac{1}{2}\sqrt{a^2+b^2} = \dfrac{1}{2}\sqrt{2 * a^2} =\\
        &= \dfrac{1}{2}\sqrt{2*(\pi_{k,1}-\pi_{k,2})^2} = \dfrac{\sqrt{2}}{2} |\pi_{k,1}-\pi_{k,2}|.
    \end{split}
\end{equation*}

The $WPD$ value between $x$ and $y$ is the average distance of all points on the warping path to the diagonal, as follows:
\begin{equation}\label{equ:wpd-d}
    WPD_d(x,y) = \dfrac{\sqrt{2}}{2.L_{\pi}}\sum_{k=1}^{L_{\pi}}|\pi_{k,1}-\pi_{k,2}|.
\end{equation}

Finally, the $WPD$ metric of a generative model is calculated, like for the $APD$ metric, between random subsets of samples from both real and generated samples:
\begin{equation}\label{equ:wpd-r}
WPD_r = \dfrac{1}{S_{wpd}}\sum_{i=1}^{S_{wpd}}WPD_d(\mathcal{S}_i,\mathcal{S}^{'}_i),
\end{equation}

\noindent where $\mathcal{S}$ and $\mathcal{S}^{'}$ are two randomly selected subsets of $\hat{\textbf{V}} = \mathcal{F}(\hat{\mathcal{X}})$, i.e., $\mathcal{S},\mathcal{S}^{'}~\subset~\hat{\textbf{V}}$ and $r~\in~\{1,2,\ldots,R\}$ is the number of repetitions of this random experiment to avoid the bias of a random selection.
The final $WPD$ metric is calculated as:
\begin{equation}\label{equ:wpd}
WPD = \dfrac{1}{R}\sum_{r=1}^{R}WPD_r ,
\end{equation}

\noindent with $WPD$ bounded between $0$ and $\dfrac{\sqrt{2}}{4}(L+1)$, and $L$ is the length of the sequences $x$ and $y$.

The characteristics of our proposed $WPD$ metric are summarized in Table~\ref{tab:summary-metrics} along with all the other metrics presented in this study.

\begin{table*}[htbp]
\centering
\caption{Summary of the Generative Models Metrics in this study.}
\label{tab:summary-metrics}
\begin{tblr}{
  width = \linewidth,
  colspec = {Q[87,c]Q[83,c]Q[50,c]Q[206]Q[187,c]Q[185,c]Q[87,c]Q[75,c]},
  hlines,
  vline{2-8} = {1}{},
  vline{1-9} = {1-11}{},
  hline{1,11} = {-}{0.18em},
}
Metric        & Category              & Space  & \SetCell[c=1]{c}Hyper-Parameters                                                                                                                                                                    & Bounds                                                                           & Interpretation                                                     & Better Version & Used in Study \\
$FID$       & fidelity              & latent & \SetCell[c=1]{c}None                                                                                                                                                                                & $0\leq FID \textless{} \infty$                   & {Higher but close\\to $FID_{real}$}                           & None           & yes           \\
$AOG$ & {fidelity/\\accuracy} & latent & \SetCell[c=1]{c}None & $0\leq AOG \leq 1$ & {Close to 1\\(100\% accuracy)} & None & yes \\
$density$   & fidelity              & latent & \labelitemi\hspace{\dimexpr\labelsep+0.5\tabcolsep}$k$: number of neighbors                                                                                                       & {$0\leq density \leq N/k$\\$N$ being the number of real samples}  & {closer to $density_{real}$\\which is close to 1}               & None           & yes           \\
$precision$ & fidelity              & latent & \labelitemi\hspace{\dimexpr\labelsep+0.5\tabcolsep}$k$: number of neighbors                                                                                                       & $0\leq precision \leq 1$                                 & closer to 1                                                        & $density$    & no            \\
$coverage$  & diversity             & latent & \labelitemi\hspace{\dimexpr\labelsep+0.5\tabcolsep}$k$: number of neighbors                                                                                                       & $0\leq coverage \leq 1$                                  & {closer to $coverage_{real}$\\which is close to $1-1/2^k$} & None           & yes           \\
$recall$    & diversity             & latent & \labelitemi\hspace{\dimexpr\labelsep+0.5\tabcolsep}$k$: number of neighbors                                                                                                       & $0\leq recall \leq 1$                                               & closer to 1                                                        & $coverage$   & no            \\
$APD$       & diversity             & latent & {\labelitemi\hspace{\dimexpr\labelsep+0.5\tabcolsep}$S_{apd}$: size of random subset\\\labelitemi\hspace{\dimexpr\labelsep+0.5\tabcolsep}$R$: number of random experiments}  & $0\leq APD < \infty$                   & Lower but close to $APD_{real}$                                 & None           & yes           \\
$ACPD$      & diversity             & latent & \labelitemi\hspace{\dimexpr\labelsep+0.5\tabcolsep}$S_{acpd}$: size of random subset\labelitemi\hspace{\dimexpr\labelsep+0.5\tabcolsep}$R$: number of random experiments & $0\leq ACPD < \infty$                             & Lower but close to $ACPD_{real}$                                & None           & yes           \\
$MMS$       & {diversity/\\novelty} & latent & \SetCell[c=1]{c}None                                                                                                                                                                                & $0\leq MMS < \infty$                   & Higher but close to $MMS_{real}$                              & None           & yes           \\
{$WPD$\\(\textbf{ours})}       & {diversity/\\warping} & raw    & {\labelitemi\hspace{\dimexpr\labelsep+0.5\tabcolsep}$S_{wpd}$: size of random subset\\\labelitemi\hspace{\dimexpr\labelsep+0.5\tabcolsep}$R$: number of random experiments}  & {$0\leq WPD \leq \dfrac{\sqrt{2}}{4}(L+1)$\\$L$ being the length of the sequence} & Depends on the application                                         & None           & yes           
\end{tblr}
\end{table*}

\section{Experimental Analysis}
\begin{figure*}[t]
    \centering
    \includegraphics[width=\textwidth]{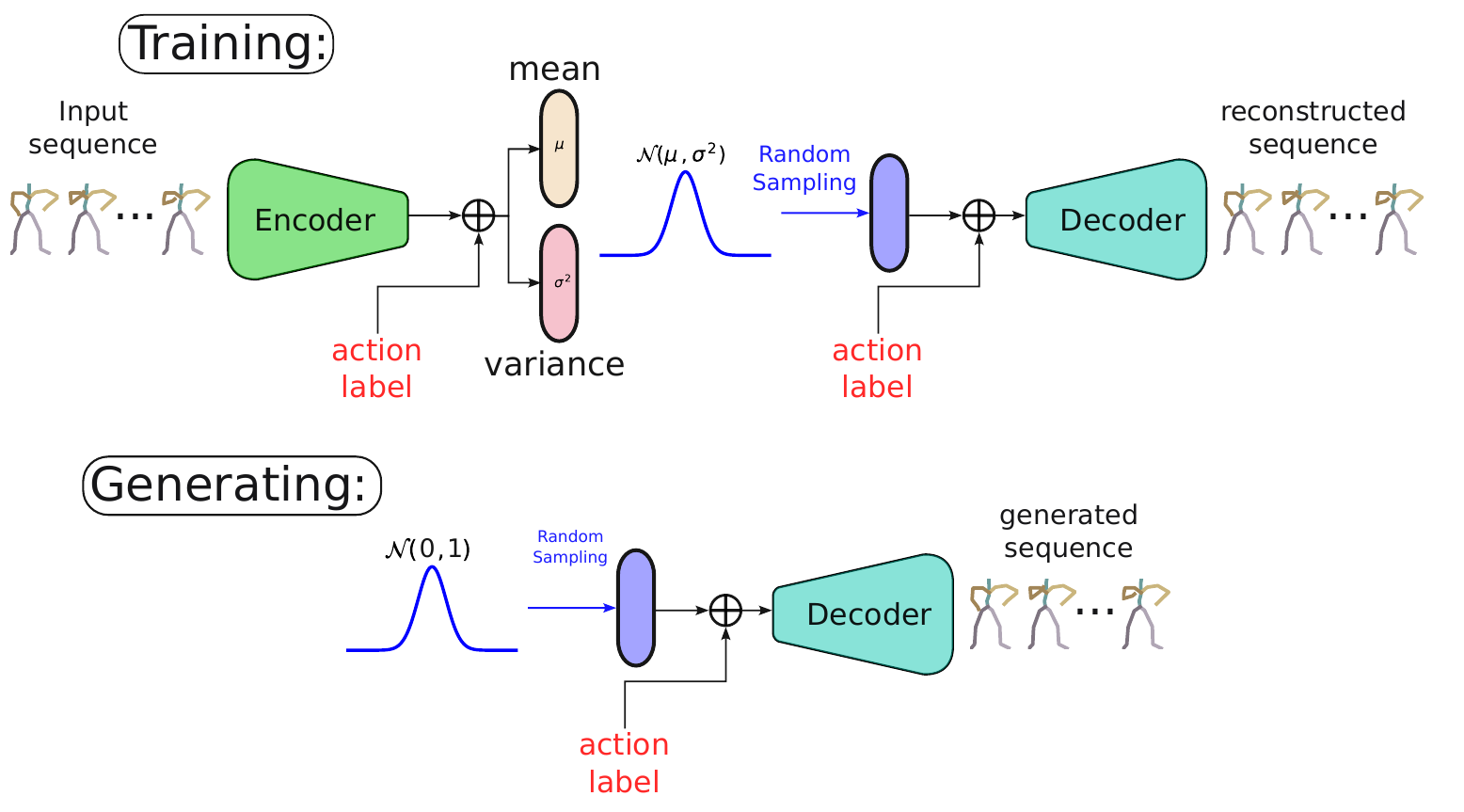}
    \caption{
    Training and generation phases of the CVAE in our experiments.
    In the training phase, both the Encoder and Decoder are trained at the same time.
    The Encoder extracts the features from the input sequences, followed by a projection into a Gaussian latent representation, while being conditioned on the input action label.
    The Decoder takes as input a random sample from the learned Gaussian space and decodes the original input sequence, while being conditioned on the input action label.
    In the generation phase, a random sample is taken from a pure Normal distribution, fed to the decoder to generate a new sequence, while being conditioned on the action label we want the generated skeleton sequence to perform.}
    \label{fig:cvae}
\end{figure*}

To quantitatively analyze each metric's behavior during evaluation, we present an experimental setup using Conditional Variational Auto-Encoders (CVAE) to generate human motion sequences.
The conditioning aspect of the VAE helps to control the generation of new examples by specifying the class of the action to generate.
A visual representation of the training/generating phases of a CVAE on human motion sequences can be seen in Figure~\ref{fig:cvae}.

\subsection{Backbone Architectures}
The CVAE model consists on an encoder-decoder architecture with a symmetrical view.
In this experiment, we choose three well known backbone neural network architectures used in the literature: (1) Convolutional Neural Networks (CNNs), (2) Recurrent Neural Networks (RNNs), and (3) Transformer Networks.

In the case of CNNs, the encoder is made of convolution blocks that contain a one-dimensional convolution layer, a batch normalization layer and an activation layer. The decoder of the CNN based CVAE contains de-convolution blocks consisting of one-dimensional convolution transpose layers, a batch normalization layer followed by an activation layer.
The number of convolution and de-convolution blocks is the same for both encoder and decoder, with the same number of filters.
The kernel sizes are chosen in a symmetrical view, i.e., if the CVAE encodes with $k_1,k_2,\ldots,k_n$ kernel sizes for each convolution block, it decodes with $k_n,\ldots,k_2,k_1$ kernel sizes for each de-convolution block.
In the rest of this work, we refer to the CNN based CVAE as CConvVAE.

In the case of RNNs, the encoder is made of Gated Recurrent Units (GRU) layers on top of each other, and outputs a vector representation.
The decoder is made of the same number of GRU layers as the encoder, with the same parameters.
However the input of the decoder is repeated as much as the number of frames in the sequence to be fed to the first GRU layer.
We refer in the rest of this work to the RNN based CVAE as CGRUVAE.

In the case of Transformer Networks, the encoder and decoder both consist of self-attention mechanisms, with the same number of layers for both and the same parameters.
We refer in the rest of this work to the Transformer Network based CVAE as CTransVAE.

\subsection{Dataset}
\begin{figure*}
    \centering
    \includegraphics[width=\textwidth]{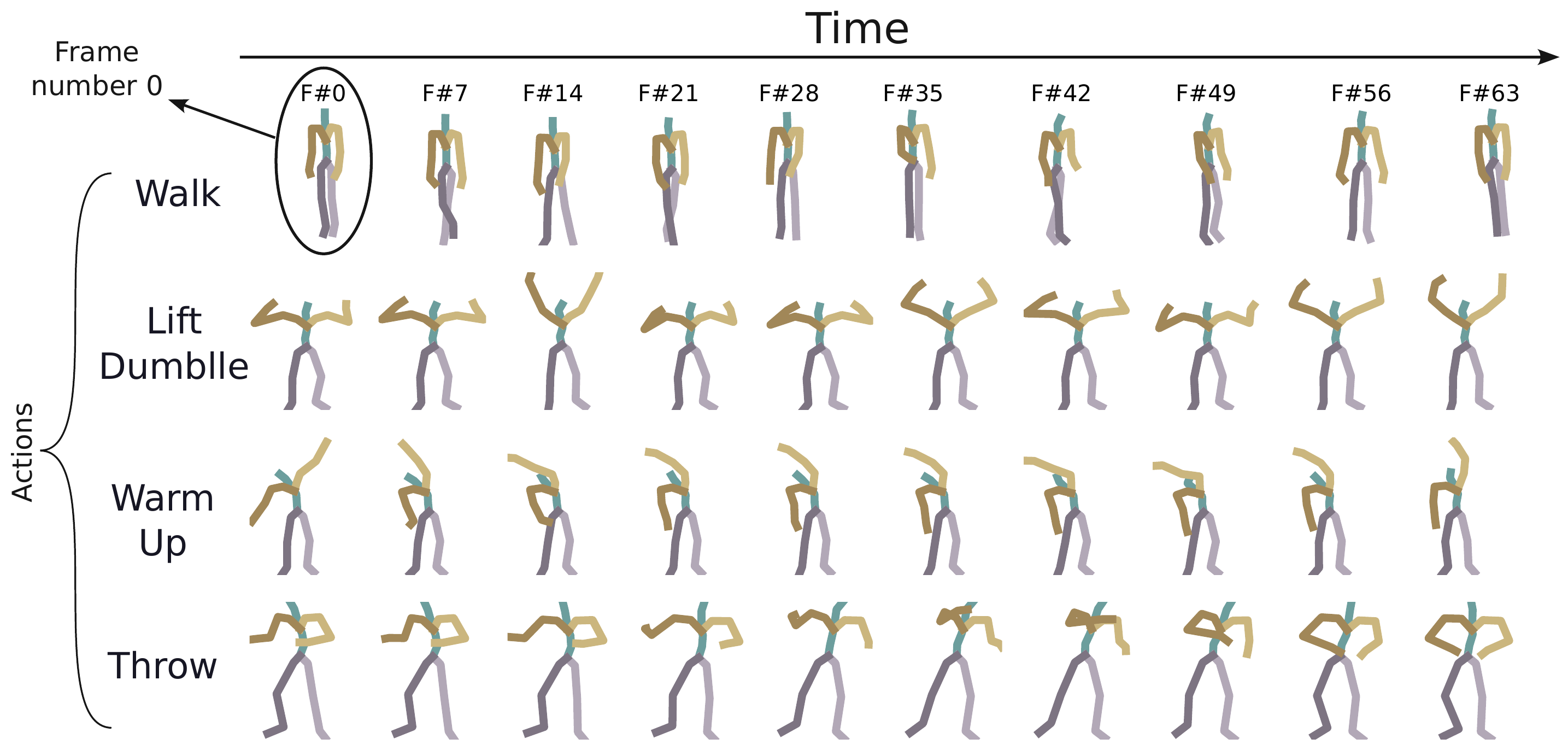}
    \caption{Visualization of some samples of the HumanAct12 dataset for four different actions: \emph{walk}, \emph{lift dumbbell}, \emph{warm up} and \emph{throw}.
    The temporal axis is the horizontal axis where we present different frames of the examples.}
    \label{fig:humanAct12}
\end{figure*}

To train the generative models, we utilize a publicly available action recognition dataset, HumanAct12.
This dataset, made available in~\citep{action2motion}, contains 3D Kinect skeleton motion sequences of subjects performing 12 actions.
These actions are: \emph{warm up, walk, run, jump, drink, lift dumbbell, sit, eat, turn steering wheel, call with phone, boxing} and \emph{throw}. 
Each of these actions contains a variety of ways of doing the action, e.g., the action \emph{warm up} contains \emph{warm up pectoral} and \emph{warm up elbow-back}.
The dataset contains $1191$ examples in total, some of which are illustrated in Figure~\ref{fig:humanAct12}.
Each example in this dataset is a sequence of skeletons, where each skeleton is made of 24 body joints with their corresponding position in the $x-y-z$ 3D space.
Moreover, each sequence can be seen as a 3D tensor of shape $(T,J,3)$, where $T$ is the length of the sequence, and $J$ is the number of joints.
These sequences can also be represented as a Multivariate Time Series of shape $(T,3.J)$, which is then used as input shape for the CVAE generative models.
Given that deep learning models require equal length sequences, we re-sample all of the samples of the HumanAct12 dataset to the average length ($75$) using the \emph{scipy.signal.resample}~\citep{2020SciPy-NMeth} functionality.
Prior to training, we normalize all sequences in the dataset using a min-max scalar on each of the $x-y-z$ dimensions independently.

It is important to note that for all the metrics used in this work, no prior train-test splits are required, instead all the dataset can be used.

\subsection{Implementation Details}
The three CVAE variants used in this work are implemented using \emph{tensorflow}~\citep{tensorflow2015-whitepaper} and trained for $2000$ epochs with a batch size of $32$.
During training, we use a learning rate decay, the ReduceLROnPlateau approach.

For the CConvVAE, we use three convolution blocks and three de-convolution blocks for the encoder-decoder architecture with $128$ filters for each layer and kernel sizes of $40,20$ and $10$ respectively (symmetrical for the decoder).

For the CGRUVAE, we use two GRU layers in the encoder.
The input vector of the decoder is repeated multiple times to form an input sequence, followed by two GRU layers.
The hidden GRU state size is set to $128$ in all of the GRU layers in both encoder and decoder.

For the CTransVAE, we use a convolution embedding at first followed by two layers of Multi-Head Attention in the encoder.
The decoder consists of two Multi-Head Attention layers followed by a de-convolution layer.
The number of filters in the convolution and de-convolution embedding layers is set to $128$ with a filter size of $40$.
The number of Attention heads is set to $4$ with a head size of $32$.

The latent space dimension of all of the three CVAE variants is set to the same value of $16$.

\subsection{Training on Different Loss Parameters}
As this experimental work aims at showcasing that a slight change in the model's parameters results in a change in the interpretation of the evaluation metrics, we choose to experiment on the model's loss parameters.
A CVAE model is trained to optimize a weighted sum of two losses, the first being is the reconstruction loss and the second being the Kullback–Leibler divergence (KL) loss.
The reconstruction loss aims to maximize the log likelihood by minimizing the mean squared error between the input and the reconstructed output.
The KL loss aims to regularize the latent space and prevents the learned Gaussian distribution from over estimating the space.
The total loss to minimize during training is as follows:
\begin{equation}\label{equ:total-loss}
\mathcal{L} = \alpha.\mathcal{L}_{rec} + \beta.\mathcal{L}_{KL},
\end{equation}

\noindent where $\alpha$ and $\beta$ are scalar weights between $0$ and $1$ for each of the reconstruction and KL loss respectively.
In the ideal case, it is preferable to maintain the following constraint:
\begin{equation}\label{equ:alpha-beta}
\alpha + \beta = 1 ,
\end{equation}

\noindent in order to have a convex sum of both losses.
From previous VAE based work in the literature, it has been seen that in most of the cases, the model converges to a more optimal state when the KL loss is assigned a smaller weight compared to the reconstruction loss.

In this work, we avoid choosing a specific couple of $(\alpha,\beta)$ and perform the experiment on different values while maintaining the constraint in Eq.~\ref{equ:alpha-beta}.
The set of values we choose for the couple $(\alpha,\beta)$ are: $(1E^{-1},9E^{-1})$, $(5E^{-1},5E^{-1})$, $(9E^{-1},1E^{-1})$, $(9.9E^{-1},1E^{-2})$, $(9.99E^{-1},1E^{-3})$, $(9.999E^{-1},1E^{-4})$.

\subsection{Class Imbalanced Generation Setup}\label{sec:setup-eval}
In this section, we propose a generation setup that respects the class imbalanced problem of the training set.
This setup is not necessarily specific to our experiment with CVAE, but with any generative model in a supervised scheme, such as human action recognition datasets.
Given that all metrics consist in comparing real and generated distribution, we believe it is important to make the evaluation as fair as possible.
When generating new samples from a generative model, it is essential to preserve the same label distribution as the one of the training set of the real samples.
The reason for this constraint is to avoid bias in the generative model for a specific sub-group of classes.
For instance, suppose a class-imbalanced dataset used to train a generative model, there is a possibility that this model is only able to generate the most populated sub-class even when being told to generate the minority.
In the case where higher number of generated samples is needed compared to the number of real samples, then we simply choose a factor to control the quantity, while preserving the label distribution.

\subsection{Results and Analysis}
In this work, we present the results on the form of a radar chart, also known as a spider chart, given the nature of our metrics having different range of values.
The usage of this kind of visualization is more feasible than simply mentioning that one model has a $0.003$ value better on a given metric for example.

For each couple of $(\alpha,\beta)$, we present one radar chart on a polygon axes, containing each four polygons: one for each of three variants of CVAE and one for the metrics computed on the real samples.
Prior to visualization, all metrics are first normalized between $0$ and $1$ independently from each other using a min max scalar.
Then, all metrics are transformed with the following formula:
\begin{equation}
    metric_{gen} = 
    \begin{cases}
        1 - (metric_{real} - metric_{gen}) & if~metric_{gen}>metric_{real}\\
        1 + (metric_{gen} - metric_{real}) & if~metric_{gen} \leq metric_{real}
    \end{cases}
\end{equation}

\noindent where $metric$ is in the set of all metrics expected for $FID$.
In the case of $FID$, the formula becomes:
\begin{equation}
    metric_{gen} = 
    \begin{cases}
        1 - (metric_{gen} - metric_{real}) & if~metric_{gen}>metric_{real}\\
        1 + (metric_{real} - metric_{gen}) & if~metric_{gen} \leq metric_{real}
    \end{cases}
\end{equation}

\noindent This transformation facilitates the reading and understanding of the radar chart.
It is based on the fact that all metrics calculated on the generative set, should be close to the metric calculated on the real set, as explained before.
The exception of the $FID$ metric is because of its nature being the only metric we are interested in decreasing it instead of increasing it.

\begin{figure*}[t]
    \centering
    \includegraphics[width=0.8\textwidth]{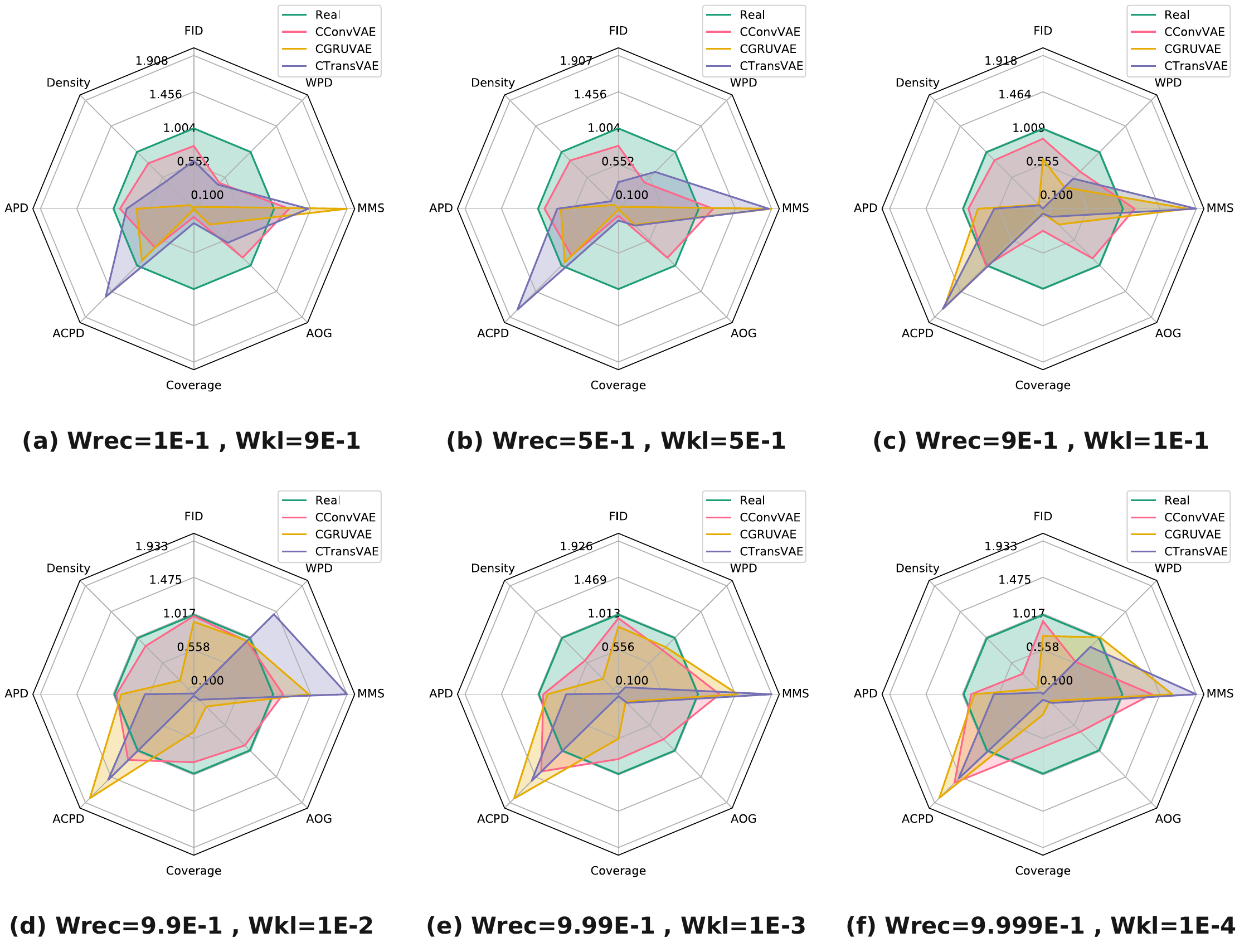}
    \caption{
    Radar charts comparing the performance of the three CVAE variants over the eight used metrics.
    For each sub-figure from a to d, we present the results of one parameter set for the three CVAE variants, by changing the values of $\alpha$ and $\beta$.
    Each radar chart contains 4 polygons, where three of these polygons are associated to each of the three CVAE variants, and the fourth polygon is for the metric values over the real set of data.
    For all metrics except for the $FID$, if the summit of polygon1 is higher than the summit of polygon2, than the metric value in reality of model1 is higher than model2.
    In the case of the $FID$, the opposite holds.}
    \label{fig:radar-charts}
\end{figure*}

In Figure~\ref{fig:radar-charts}, we present the radar charts for all pairs of $(\alpha,\beta)$ for the three variant CVAE and the real set.
For each metric, a summit of a generative model's polygon being lower than the summit of the real polygon means that $metric_{gen} < metric_{real}$, except for the $FID$ metric it signifies that $FID_{gen} > FID_{real}$.

In the case where an application is interested in being most optimal on all metrics, then the generative model's polygon should be most identical to the real polygon, except for the $MMS$ metric summit.
It is important to recall that for the $MMS$ metric, $MMS_{gen}$ should be higher to a certain limit than $MMS_{real}$, which is interpreted visually by the summit of the generative model's polygon being higher than the summit of the real polygon at the $MMS$.

Figure~\ref{fig:radar-charts} highlights that it may be difficult to find a generative model winning on all metrics simultaneously, as a change in the backbone architecture or in the loss parameters may result in a significant change in the value of some metrics.
According to our experimental evaluation on the HumanAct12 dataset, the only case scenario where it may be possible to choose a best model, CConvVAE, is for a very specific set of values for the pair $(\alpha,\beta)$, corresponding to Figure~\ref{fig:radar-charts}-d.
However, this kind of parameter search is not always feasible. This is why finding the \emph{best} model on all possible metrics can be a difficult task.
Instead, depending on the target application, we should expect to search for a \emph{best} model on some specific set of, or even only one, metric.

In what follows, we present an interpretation of the results by comparing the three CVAE on each metric individually. For each metric, we detail the interpretation of a model being \emph{best} on this metric.

\subsubsection{$FID$}
For some pairs of $(\alpha,\beta)$, it can be seen in Figure~\ref{fig:radar-charts} that CConvVAE has the smallest $FID$ value that is still higher but close to the $FID_{real}$.
This means that the generated samples produced by CConvVAE are more reliable in terms of fidelity compared to CGRUVAE and CTransVAE.
The CConvVAE model is able to learn a distribution \pg~that requires less energy to transform into \pr~compared to the distributions \pg~learned by CGRUVAE and CTransVAE.
However in some cases, even being the variant with the smallest and closest value, the gap is large (Figures~\ref{fig:radar-charts}-a-b-c).

\subsubsection{Density}
Similar to the interpretation of the $FID$ metric, for some pairs of $(\alpha,\beta)$, the CConvVAE is the generative model with a $Density$ value closer to the $Density_{real}$ compared to the two other CVAE variants.
This implies that the likelihood of samples, generated by \pg, being generated by \pr in the case of CConvVAE, is higher compared to that in the cases of CGRUVAE and CTransVAE.
However, similar to the $FID$ metric, even if in some cases the CConvVAE has the closest value to $Density_{real}$ a significant gap can be observed between $Density_{gen}$ and $Density_{real}$.

\subsubsection{AOG}
The $AOG$ metric is essential to assess if the generative model's conditional aspect is effective or not.
This conditional mechanism may not be always efficient, while training generative models because of lack of data, underfitting or hyper-parameters settings.
For instance, it can be seen from Figure~\ref{fig:radar-charts}, that even though CConvVAE is the winning model on this metric for all parameters settings, it would not mean that the conditional mechanism is not failing.
This can be seen in Figure~\ref{fig:radar-charts}-f where the $AOG$ value in the case of CConvVAE is far fro the $AOG_{real}$, which means the conditional mechanism is failing in this hyper-parameter settings.
In the case of CTransVAE and CGRUVAE the mechanism seems to fail for any hyper-parameter setting, which means the generative model is not able to learn how to control the sub-class of each sample being used to learn the model's hyper-parameters.

For all of the three above fidelity based metrics, the CConvVAE is considered most reliable compared to other CVAE variants.

\subsubsection{APD}
On the one hand, in terms of the $APD$ diversity metric, the CConvVAE triumphs over this metric 
On the one hand, the CConvVAE seems more diverse according to the $APD$ metric by having the closest value to the $APD_{real}$ on all hyper-parameters settings with an almost ideal $APD$ value for Figure~\ref{fig:radar-charts}-d.
On the other hand, the CGRUVAE performs almost as good as the CConvVAE on the $APD$ metric in some of the parameters settings.
This showcases that even though CConvVAE can be significantly better than CGRUVAE on one fidelity metric such as $FID$, it would not mean that CGRUVAE cannot perform well on other metrics.

\subsubsection{ACPD}
The $ACPD$ metric evaluates the space diversity per sub-class of \pr and \pg~on contrary to the $APD$ metric, which evaluates space diversity in general over all classes.
From Figure~\ref{fig:radar-charts} it can be seen that in almost half the cases of hyper-parameters settings, all three CVAE variants perform better on the $ACPD$ metric compared to the real data ($ACPD_{real}$).
This indicates that the generated space is more diverse per sub-class compared to the real space.
This last interpretation is not entirely correct, as this phenomena where $ACPD_{gen} > ACPD_{real}$ is mostly a result of overfitting, instability and class imbalanced issue during training of the generative model.
In addition, winning on one metric does not signifies winning on all metrics, for instance even though CGRUVAE outperforms CConvVAE in Figure~\ref{fig:radar-charts}-a in terms of $ACPD$, it would not directly conclude that CGRUVAE is more diverse per sub-class than CConvVAE.
This is due to the fact that on the same hyper-parameter settings, the $FID$ value of CGRUVAE is much higher than $FID_{real}$ resulting in almost random generation, for which the $ACPD$ values are not reliable.
This highlights the importance of using different evaluation metrics.

\subsubsection{Coverage}
For most hyper-parameter settings, the CConvVAE outperforms other CVAE variants in terms of $coverage$.
This signifies that in the case of CConvVAE more samples generated by \pr~can also be generated by the learned distribution \pg~compared to the case of CGRUVAE and CTransVAE.
However, in the cases of Figures~\ref{fig:radar-charts}-a and b, the CConvVAE does not outperform other variants on the $coverage$ metric, instead CTransVAE wins on the $coverage$ metric.
This means, even though for the same hyper-parameters CConvVAE performs well in terms of $APD$ diversity, it performs very badly in terms of $coverage$.

This raises the question: what is the difference between $APD$  and $coverage$ if they both quantify diversity? 
While both metrics rely on a latent space view with distance based measure (Euclidean distance), they differ in terms of quantification.
On the one hand the $APD$ metric relies on distance value between a random selection of pairs, 
quantifying the volume of space taken by \pr~and \pg.
On the other hand, the $coverage$ metric relies on the nearest neighbor portions between real and generated samples, 
quantifying 
how much volume do the generated samples take in \pr.

\subsubsection{MMS}
The $MMS$ metric evaluates the diversity in terms of novelty of the generated samples, where the expected value of $MMS_{gen}$ should be higher but still close to $MMS_{real}$.
This makes the interpretation of this metric a bit more difficult to the other metrics.
In the case of CConvVAE, the $MMS$ values seem to be the most stable between all three variants, as it is higher but still close to $MMS_{real}$, on contrary to CTransVAE, where $MMS$ takes over the limit of the radar plot. This suggests that the CConvVAE model is able to generate more novel human motion sequences.

\subsubsection{WPD}
The $WPD$ metric evaluates the diversity from a temporal perspective, as it evaluates how much warping and temporal distortions do exist between samples.
For the interpretation of $WPD$, three cases do exist:
\begin{itemize}
    \item Case 1: $|WPD_{gen}-WPD_{real}| < \epsilon$ for $\epsilon <<< 1$, which indicates a perfect re-creation of all temporal distortions in \pg provided by \pr.
    \item Case 2: $WPD_{gen} >>> WPD_{real}$, which indicates the generative model found that some temporal distortions exist in \pr and learned how to create similar but new ones.
    \item Case 3: $WPD_{gen} <<< WPD_{real}$, which is an indication of the model not being able to re-create temporal distortions and is rather always generating the same type of temporal distortions.
\end{itemize}
From Figure~\ref{fig:radar-charts}, it can be seen that in all cases, except for CTransVAE in Figure~\ref{fig:radar-charts}-d, all $WPD$ values are smaller than $WPD_{real}$.
This indicates that in most cases the models can re-create some of the temporal distortions but not all of them.
However, in one of the parameters settings, Figure~\ref{fig:radar-charts}-d, the gap between $WPD_{real}$ and $WPD$ for CConvVAE and CGRUVAE is very small, which indicates a perfect re-creation of the temporal distortions in \pr.

\subsection{Choosing The Right Metric}

In overall, our findings showcase that while the CConvVae model seems to be more appropriate when trained on the HumanAct12 dataset, some small changes in hyper-parameters can significantly affect its performance on most of the metrics.
This suggests that it is not always easy to find the \emph{best model} to win over all metrics, and we believe that it should always depend on the target application.
For instance, if the workflow of generating new human motion sequences is dedicated for a gaming setup, fidelity metrics such as $FID$ are useful
but not as important as diversity metrics.
In this scenario, generated samples are usually foreseen to have a very high number of possibilities of doing an action depending on the gaming environment of a user.
Moreover, if the generation task is in the aspect of medical research such as generating rehabilitation movements for patients, than fidelity plays a huge role in this case as we want to show the patient an exact replica of how a human patient should do the exercise.

\subsection{Reproducibility}

In order to favor the reproducibility of our work, we publicly release our code, available at \texttt{https://github.com/MSD-IRIMAS/Evaluating-HMG}. We believe this repository is easily manageable and user-friendly.
It allows a user to evaluate an existing or a new generative model on all the considered metrics using the same classifier pre-trained on the HumanAct12 dataset.
The code provides different options from which a user can choose the metrics to compute as well choosing the set of parameters of each metric.

\section{Conclusions}
In this work, we presented a comprehensive review on evaluation metrics used to quantify reliability of generative models for the task of human motion generation.
By leveraging the fact that human motion data are temporal data represented as multivariate time series, we also proposed a novel metric to assess diversity in terms of temporal distortion.
Given the set of all the considered metrics, we proposed a unified evaluation setup that enables a fair comparison between different models.
Through some experiments with three variants of generative models and a publicly available dataset, our experiments highlighted that it is not straightforward to find \emph{The One Metric To Rule Them All}.
Instead, our findings showcased that a set of different metrics is often more appropriate to conclude how reliable a model is compared to other models.
Our work is supported by a user-friendly publicly available code that can be used to calculate all of the used metrics on any generative model with any possible parametrization.


We believe this work can be a good starting point for new comers into the field of human motion generation and can give a clear definition to a unified evaluation setup.
However, we do not claim that the presented metrics in this paper are the only one available, as this research field is exponentially expanding.
It is also important to note that most metrics suppose availability of labels in the real data, however this is not always the case and metrics in this case should be adapted.

\section*{Acknowledgments}
This work was supported by the ANR DELEGATION project (grant ANR-21-CE23-0014) of the French Agence Nationale de la Recherche.
The authors would like to acknowledge the High Performance Computing Center of the University of Strasbourg for supporting this work by providing scientific support and access to computing resources.
Part of the computing resources were funded by the Equipex Equip@Meso project (Programme Investissements d’Avenir) and the CPER Alsacalcul/Big Data.
The authors would also like to thank the creators and providers of the HumanAct12 dataset.

\bibliographystyle{model2-names}
\bibliography{refs}



\end{document}